\documentclass[11pt]{article}

\usepackage[preprint]{acl}

\usepackage{times}
\usepackage{latexsym}
\usepackage{pifont}
\usepackage[T1]{fontenc}

\usepackage[utf8]{inputenc}

\usepackage{microtype}
\usepackage{inconsolata}
\usepackage{array}

\usepackage[table]{xcolor} 
\usepackage{multirow}

\usepackage{amsmath}
\usepackage{amssymb}
\usepackage{graphicx}
\usepackage{enumitem}
\usepackage{booktabs}
\usepackage{subcaption}
\usepackage{makecell}
\usepackage{pifont} 
\usepackage[export]{adjustbox} 

\definecolor{HeaderColor}{RGB}{235, 242, 250} 
\definecolor{DataColor}{RGB}{253, 253, 253}   
\definecolor{HighlightColor}{RGB}{255, 248, 240}
\definecolor{lightcyan}{rgb}{0.88, 0.95, 1.0}

\newcommand{\best}[1]{\cellcolor{red!20}#1}    
\newcommand{\second}[1]{\cellcolor{blue!20}#1} 
\begin{document}
\title{DIVER: Dynamic Iterative Visual Evidence Reasoning for \\ Multimodal Fake News Detection}

\author{
  Weilin Zhou$^{1,2,\ast}$, 
  Zonghao Ying$^{3,\ast}$, 
  Chunlei Meng$^{4}$, 
  Jiahui Liu$^{5}$, 
  Hengyang Zhou$^{6}$, \\
  \textbf{Quanchen Zou}$^{2,\dagger}$, 
  \textbf{Deyue Zhang}$^{2}$, 
  \textbf{Dongdong Yang}$^{2}$, 
  \textbf{Xiangzheng Zhang}$^{2}$ \\
  $^1$Xinjiang University \quad $^2$360 AI Security Lab \quad $^3$Beihang University \\
  $^4$Fudan University \quad $^5$Central South University \quad $^6$Nanjing University \\
  \small $^\ast$Equal contribution \quad $^\dagger$Corresponding author \\
}

\maketitle

\begingroup
\renewcommand\thefootnote{}
\footnote{* Equal contribution.}
\footnote{$\dagger$ Corresponding author: Quanchen Zou.}
\footnote{The work was done at 360 AI Security Lab.}
\addtocounter{footnote}{-3}
\endgroup

\begin{abstract}
Multimodal fake news detection is crucial for mitigating adversarial misinformation. Existing methods, relying on static fusion or LLMs, face computational redundancy and hallucination risks due to weak visual foundations. To address this, we propose DIVER (Dynamic Iterative Visual Evidence Reasoning), a framework grounded in a progressive, evidence-driven reasoning paradigm. DIVER first establishes a strong text-based baseline through language analysis, leveraging intra-modal consistency to filter unreliable or hallucinated claims. Only when textual evidence is insufficient does the framework introduce visual information, where inter-modal alignment verification adaptively determines whether deeper visual inspection is necessary. For samples exhibiting significant cross-modal semantic discrepancies, DIVER selectively invokes fine-grained visual tools (e.g., OCR and dense captioning) to extract task-relevant evidence, which is iteratively aggregated via uncertainty-aware fusion to refine multimodal reasoning. Experiments on Weibo, Weibo21, and GossipCop demonstrate that DIVER outperforms state-of-the-art baselines by an average of 2.72\%, while optimizing inference efficiency with a reduced latency of 4.12 s.
\end{abstract}

\section{Introduction}

The evolution of multimodal fake news has become increasingly adversarial, with fabricators pairing misleading text with contextually ambiguous imagery to complicate verification. While early studies focused on unimodal cues~\citep{chakraborty2022detecting, ni2021mvan}, recent efforts have shifted toward multimodal fake news detection (MFND), where effective reasoning over cross-modal evidence is essential.

\begin{figure}[t]
    \centering
    \begin{subfigure}{\columnwidth}
        \centering
        \begin{minipage}[c]{0.05\columnwidth}
            (a)
        \end{minipage}%
        \begin{minipage}[c]{0.93\columnwidth}
            \centering
            \includegraphics[width=0.9\linewidth]{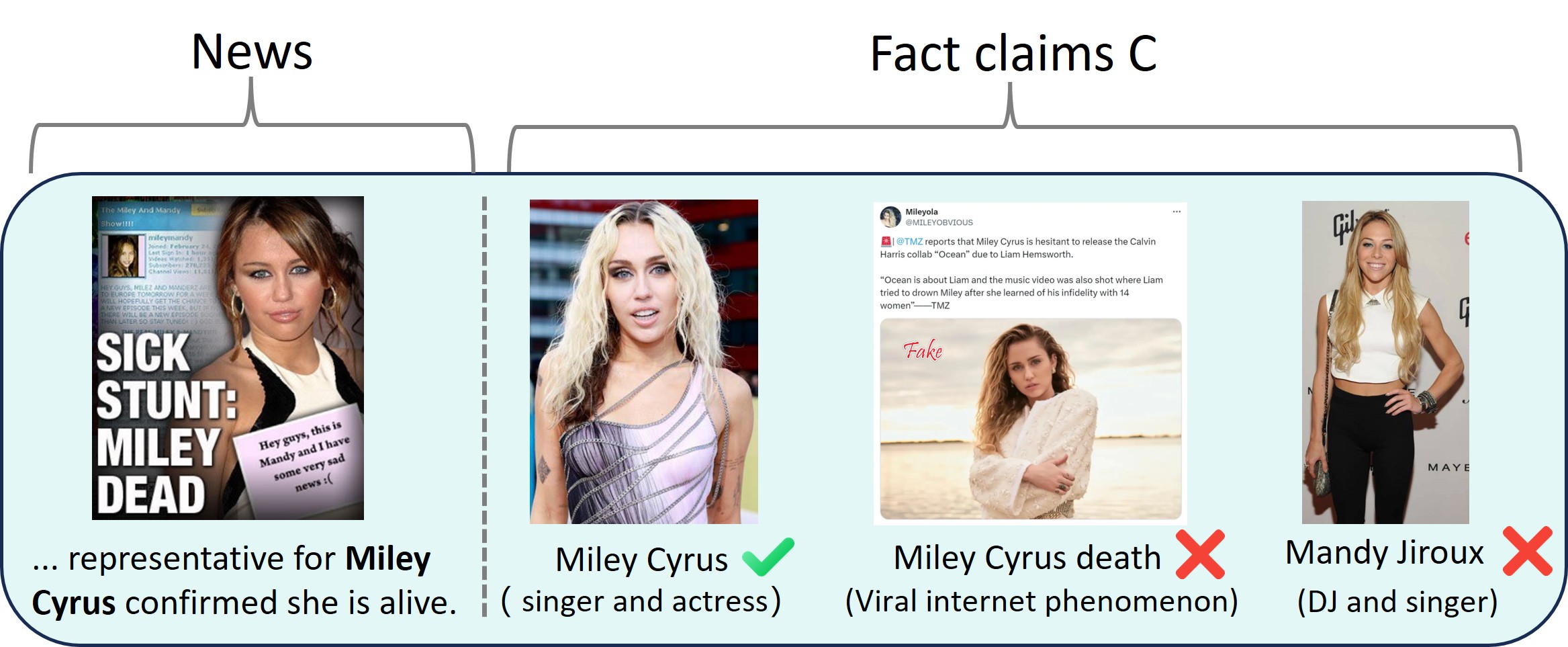} 
        \end{minipage}
        \label{fig:sub_a}
    \end{subfigure}

    \vspace{0.2cm} 

    \begin{subfigure}{\columnwidth}
        \centering
        \begin{minipage}[c]{0.05\columnwidth}
            (b)
        \end{minipage}%
        \begin{minipage}[c]{0.93\columnwidth}
            \centering
            \includegraphics[width=0.9\linewidth]{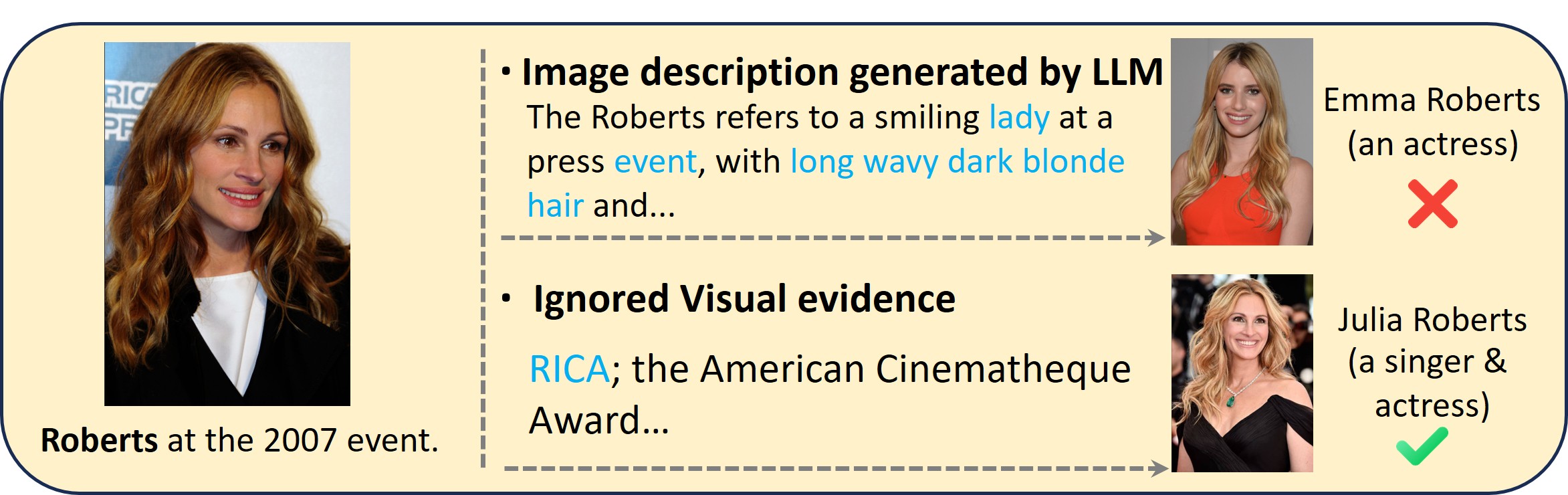}
        \end{minipage}
        \label{fig:sub_b}
    \end{subfigure}

    \caption{Two primary limitations of existing multi-modal methods: (a) noise induced factual ambiguity, and (b) visual hallucination due to weak grounding.}
    \label{fig:framework_comparison_vertical}
\end{figure}

Recent MFND frameworks increasingly rely on large language models (LLMs) to enhance multimodal reasoning, such as INSIDE \cite{wang2025bridging} and LIFE \cite{wang2025prompt}. Fundamentally, these approaches are static because they indiscriminately apply the same depth of visual processing to all inputs, failing to distinguish between obvious truths and subtle fabrications. Conversely, DIVER operates on a dynamic paradigm: it functions as an active decision-maker that regulates reasoning depth, triggering iterative visual evidence gathering only when high-level semantic alignment fails.Although leveraging LLMs for inference, these methods operate through single-pass interactions with global visual embeddings. This coarse representation fails to bridge the semantic gap between abstract claims and concrete visual facts. Without explicit verification anchored in discrete visual evidence, LLMs are prone to hallucination when subtle visual details contradict textual narratives, leading to the validation of false claims based on unsupported contexts.

In contrast, traditional multimodal fusion approaches~\citep{zhu2025ken, ying2023bootstrapping} adopt static processing pipelines that indiscriminately incorporate deep visual features for all samples. While exhaustive visual forensics can improve robustness, it introduces two inherent limitations. First, mandatory deep visual analysis incurs substantial computational overhead, which is redundant for samples that can be resolved through shallow semantic alignment. Second, forcibly injecting complex visual features when textual evidence is already sufficient may introduce irrelevant noise, potentially degrading detection accuracy. These limitations reveal a fundamental mismatch between reasoning depth and computational efficiency in existing MFND systems.

To address this mismatch, MFND requires a dynamic mechanism that regulates reasoning depth based on cross-modal necessity. Specifically, visual forensics should be selectively activated only when semantic discrepancies between text and image indicate potential deception, rather than being uniformly applied to all inputs.

Based on this principle, we propose DIVER (Dynamic Iterative Visual Evidence Reasoning), which formulates multimodal fake news detection as a dynamic decision-making process. DIVER first establishes a strong text-based baseline to assess intra-modal consistency and eliminate unreliable claims. It then evaluates coarse cross-modal alignment and utilizes an inter-modal alignment score to quantify the semantic consistency between textual claims and visual representations. When this alignment score falls below a learned threshold (signaling high semantic discrepancy), the framework transitions from fast screening to deliberate visual reasoning. This process selectively invokes fine-grained forensic tools to extract explicit evidence for verification, reflecting a dual-stage reasoning structure analogous to System 1 and System 2 processes.

This alignment-guided gating strategy enables DIVER to effectively balance accuracy and efficiency. Across three benchmarks, DIVER achieves 91.6\% accuracy on GossipCop and 94.8\% on Weibo, while reducing average inference latency to 4.12s. In comparison, representative baselines such as LIFE and INSIDE incur higher latencies of 6.83s and 4.83s, respectively. By concentrating computational resources on samples exhibiting significant inter-modal inconsistencies, DIVER accelerates inference while mitigating hallucinations caused by irrelevant visual details.

Our contributions are summarized as follows:
\begin{itemize}
    \item We formulate multimodal fake news detection as a dynamic decision-making process that prioritizes linguistic verification and selectively activates visual reasoning only when cross-modal discrepancies arise.
    \item We propose an alignment-guided gating mechanism that dynamically adjusts the granularity of analysis, enabling efficient filtering of low-risk samples while selectively invoking fine-grained visual forensics (e.g., OCR and dense captioning) for ambiguous cases.
    \item Extensive experiments across multiple benchmarks demonstrate that DIVER consistently outperforms state-of-the-art methods while achieving superior inference efficiency.
\end{itemize}

\section{RELATED WORK}
\subsection{Multimodal Fake News Detection} Existing methodologies fall primarily into two paradigms: unimodal methods and cross-domain generalization frameworks.

Early unimodal methods \cite{liu2024fakenewsgpt4,jiang2025cross} failed to detect sophisticated cross-modal contradictions, promoting multimodal detection. However, current frameworks \cite{zhou2025robustrealiblemultimodalfake,zhu2025ken,liu2025modality} often employ static fusion on global embeddings, interpreting images superficially.

These approaches face two limitations: (1) Weak visual grounding: Overlooking fine-grained details causes semantic gaps between textual claims and visual evidence. We address this by activating visual forensics only upon cross-modal discrepancies. (2) Noise injection: Indiscriminate fusion dilutes accuracy with irrelevant information~\cite{jin2017multimodal}. We mitigate this via a dynamic strategy regulating reasoning depth, selectively invoking visual analysis to avoid exhaustive processing.

\subsection{Application of LLM in Fake News Detection} 
Multimodal LLMs enhance fake news detection \cite{10843779,tian2025symbolic} by leveraging generative reasoning for active evidence extraction \cite{chen2025lvagent}. Acting as forensic analysts, these models decompose complex narratives into atomic factual claims \cite{farhangian2025dres} and generate interpretable visual descriptors, such as OCR, to bridge cross-modal semantic gaps.

Furthermore, LLMs facilitate consistency checking via reflection, explicitly reasoning about contradictions between claims and multimodal evidence \cite{chalehchaleh2025addressing}. While susceptible to fabricating correlations without rigorous fine-grained visual grounding, integrating LLMs as dynamic engines within multi-stage frameworks \cite{chalehchaleh2025addressing,bu2025enhancing} remains a robust strategy for mitigating hallucinations.

\section{Method}
We propose DIVER to overcome static fusion limitations. Inspired by the dual process theory—where initial perceptions are reassessed upon contradiction, DIVER operates as a dynamic decision-making pipeline rather than a static cascade.

\begin{figure*}[t] 
    \centering
    \includegraphics[width=\textwidth]{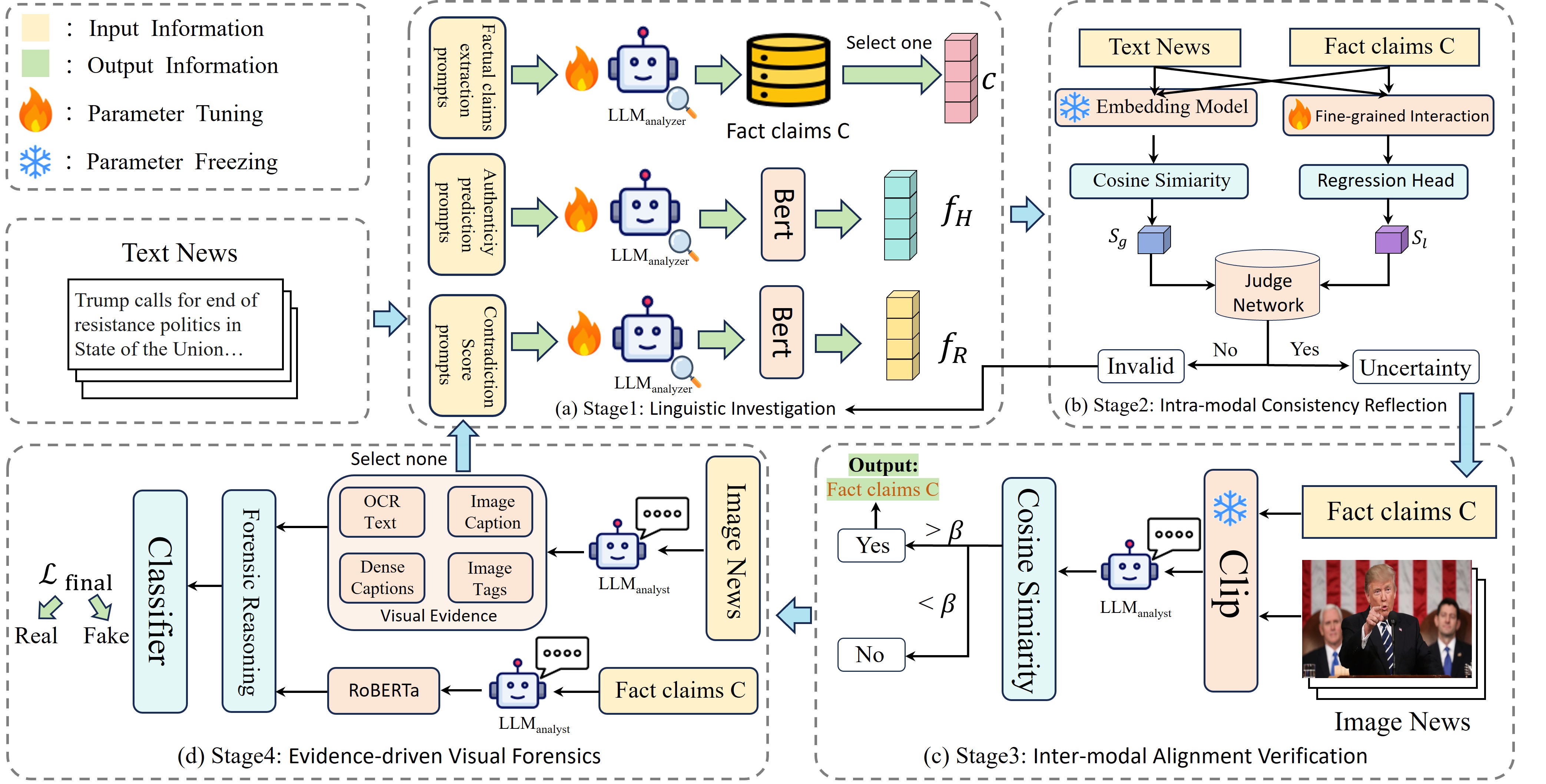} 
    \caption{
    Overview of the DIVER framework featuring two explicit feedback loops. The pipeline initiates with linguistic investigation to verify intra-modal consistency. Subsequently, a cross-modal alignment gate dynamically regulates reasoning depth: samples exhibiting high semantic consensus are resolved immediately, while disjoint cases trigger evidence-driven visual forensics, selectively activating visual analysis capabilities (e.g., OCR) to aggregate evidence for claim refinement.
    }
    \label{fig:model_architecture}
\end{figure*}

Figure \ref{fig:model_architecture} illustrates the framework's four progressive stages. Initially, the process extracts atomic statements and verifies intra-modal consistency. Verified statements are then compared with visual content via alignment gates. High consistency samples are processed immediately for efficiency, while discrepancies trigger evidence-driven visual forensics. Here, visual evidence is iteratively aggregated to refine textual statements, ensuring a rigorous System 2 reasoning process.

\subsection{Linguistic Investigation}
This stage establishes a robust linguistic baseline from the raw news text $T$. 
Unlike previous methods that treat text as a monolithic feature, we employ a fine-tuned $\text{LLM}_{analyst}$ to decouple explicit factual content from implicit stylistic cues via three parallel tasks. This parallel design ensures the independent acquisition of diverse linguistic features, preventing the entanglement of subjective stylistic patterns with objective factual claims:

Factual Claims Extraction ($P_{extract}$): To isolate verifiable information from noise, the LLM decomposes $T$ into a set of atomic, independently verifiable fact claims $C = \{c_1, c_2, ..., c_k\}$. This discrete representation facilitates precise cross-modal alignment in subsequent stages.

Linguistic Feature Analysis: Simultaneously, the LLM scrutinizes $T$ for common linguistic patterns in fake news, such as emotional manipulation, sensationalism, and logical fallacies. The generated natural language analyses are then encoded by a pretrained BERT into dense feature vectors: $f_H$ (capturing authenticity/style) and $f_R$ (capturing logical contradiction).

Formally, the linguistic baseline is constructed as:
\begin{align}
    C &= \text{LLM}_{\text{analyst}}(P_{\text{extract}}, T), \\
    f_H &= \text{BERT}(\text{LLM}_{\text{analyst}}(P_{\text{auth}}, T)), \\
    f_R &= \text{BERT}(\text{LLM}_{\text{analyst}}(P_{\text{contra}}, T)).
\end{align}

\subsection{Intra-modal Consistency Reflection}
To prevent error propagation caused by LLM hallucinations, where the extracted claims $C$ might deviate from the original semantics of $T$, we introduce a verification module to assess faithfulness before any cross-modal interaction.

Dual Branch Verification. We quantify the semantic consistency between $T$ and $C$ from both global and local perspectives:
\begin{itemize}
    \item \textbf{Global Consistency ($S_g$):} We utilize the SFR-Embedding-Mistral \cite{meng2024sfrembedding} to compute the cosine similarity of sentence level embeddings between $T$ and $C$, ensuring the overall semantic gist is preserved.
    \item \textbf{Fine-grained Interaction ($S_l$):} A trainable cross-attention network is employed to capture subtle entity level discrepancies or omissions that global embeddings might overlook.
\end{itemize}

A lightweight judge network, implemented as a binary classifier, aggregates the consistency metrics $[S_g, S_l]$ to determine the quality of $C$.

Self-Correction Mechanism.
Addressing the stochastic nature of generative models, we propose a robust retry mechanism. If judgenet classifies the extraction as invalid, a self-correction loop is triggered. The model feeds the error signal back to the LLM to regenerate claims:
\begin{equation}
C^{(t+1)} = \text{LLM}_{analyst}(P_{correct}, T, C^{(t)}, \text{ErrorMsg}),
\end{equation}
where $\text{ErrorMsg}$ represents a diagnostic natural language cue based on inconsistency metrics, explicitly instructing the LLM to correct semantic discrepancies in subsequent iterations.

If the output remains invalid after $\tau$ attempts, we activate a fallback strategy where $C$ is replaced by a standard summarization of $T$. This ensures pipeline completeness and prevents hallucinations by avoiding forced details from ambiguous text. Instead, summarization serves as a conservative semantic anchor, preserving the global gist without introducing unsupported atomic claims. Only claims verified as semantically faithful proceed to the next stage.

\subsection{Inter-modal Alignment Verification}
For verified claims $C$, we assess their compatibility with the visual modality $I$ using a frozen CLIP model. This stage serves as a lightweight, early-stage filtering mechanism. Formally, we compute a cross-modal alignment score $S_{inter}$ as:
\begin{equation}
    S_{inter} = \text{CLIP}_{\text{text}}(C) \cdot \text{CLIP}_{\text{img}}(I)^{\top}
\end{equation}

We utilize a learnable threshold $\beta$ to dynamically determine the necessity of deep forensic analysis. The decision logic is formulated as:
\begin{equation}
    \label{eq:gate}
    \text{Action} = 
    \begin{cases} 
    \text{Proceed to Stage 4}, & \text{if } S_{inter} < \beta \\
    \text{Skip Forensics}, & \text{if } S_{inter} \ge \beta 
    \end{cases}
\end{equation}
where $S_{inter} < \beta$ indicates a significant cross-modal discrepancy or insufficient grounding, signaling that the visual content fails to effectively support the textual claims. In such cases, the low alignment necessitates external evidence from the subsequent visual forensics stage. Conversely, $S_{inter} \ge \beta$ implies sufficient semantic alignment (or that the image is generic), allowing the model to bypass the computationally expensive Stage 4.

\subsection{Evidence-driven Visual Forensics}
We define evidence-driven as an active paradigm where the model explicitly seeks and verifies discrete visual artifacts to adjudicate textual claims, bypassing passive multimodal fusion. Activated by the inter-modal gate, this stage instantiates a deliberate System~2 process to resolve fine-grained semantic conflicts. Instead of exhaustive signal processing, we perform targeted visual probing to extract task-relevant evidence $E_v$ conditioned on textual claims.

\indent \textbf{Adaptive Visual Evidence Probing.}
Given verified claims $C$, we select specialized extractors based on semantic categories. Claims involving explicit attributes (e.g., names, timestamps) prioritize optical character recognition; activities or interactions invoke dense descriptive representations; and entity-centric claims rely on object-level analysis. This routing avoids redundancy while ensuring contradictions are examined using the most informative visual cues.

\indent \textbf{Conditional Recursive Refinement.}
Cheapfakes often exploit surface-level plausibility despite visual incompatibility. We introduce a conditional feedback mechanism, computing a refutation score between claims $C$ and evidence $E_v$:
\begin{equation}
S_{\text{refute}} = \text{RoBERTa}(C) \cdot \text{RoBERTa}(E_v)^\top.
\end{equation}

A rollback to Stage~1 is triggered if $S_{\text{refute}} < \gamma$ (a strict contradiction threshold), indicating visual evidence violates $C$. In such cases, $E_v$ is injected as grounded context to re-instantiate claims:
\begin{equation}
C_{\text{refined}} = \text{LLM}_{\text{analyst}}(P_{\text{re-extract}}, T, E_v).
\end{equation}

This recursive step enables revising ambiguous interpretations by anchoring reasoning in concrete visual artifacts, preventing error propagation.

\indent \textbf{Semantic Evidence Consolidation.}
Upon convergence, visual evidence $E_v$ is integrated semantically rather than as raw features. The LLM distills heterogeneous findings into a compact summary, encoded as a unified representation $f_v$. When Stage~4 is skipped by the gating mechanism (Eq.~\ref{eq:gate}), $f_v$ is set to a zero vector with mask $m_v = 0$; otherwise, $m_v = 1$. This formulation enforces representational alignment and ensures visual evidence influences reasoning only when explicitly activated.

\subsection{Uncertainty-Aware Dynamic Fusion}
\label{sec:fusion}

We propose an Uncertainty-Aware Dynamic Fusion (UDF) mechanism to effectively aggregate features $F = \{f_{t}, f_H, f_R, f_v\}$ despite the intermittent missing modalities caused by the skipping mechanism.

First, we explicitly derive a refined feature $\hat{f}_k$ for each branch weighted by its estimated reliability $\mu_k$:
\begin{equation}
\mu_k = \sigma(\mathbf{w}_\mu^T \tanh(\mathbf{W}_p f_k + \mathbf{b}_p)), \quad \hat{f}_k = \mu_k \cdot f_k.
\label{eq:confidence}
\end{equation}
where $\sigma$ denotes the sigmoid function and $\mathbf{W}_p$ projects features into a latent confidence space.

Conventional fusion methods misinterpret zero-padded features from bypassed stages as signals, injecting noise. To resolve this, UDF employs a modified masked attention mechanism \cite{vaswani2017attention} using binary masks $M = \{m_{t}, m_H, m_R, m_v\}$. We introduce a learnable task query $\mathbf{q}_{task}$ to guide the attention distribution:
\begin{equation}
e_k = \frac{(\mathbf{W}_Q \mathbf{q}_{task})^T (\mathbf{W}_K \hat{f}_k)}{\sqrt{d_h}},
\end{equation}
\begin{equation}
\alpha_k = \text{Softmax}(e_k + \mathcal{M}_k).
\end{equation}
where $\mathcal{M}_k = -\infty$ if $m_k=0$ (skipped) and $0$ otherwise. This masking ensures that missing modalities (e.g., $f_v=\mathbf{0}$) contribute exactly zero influence to the probability distribution, maintaining mathematical rigor.

The final representation is computed as:
\begin{equation}
    f_{\text{final}} = \sum_{k} \alpha_k (\mathbf{W}_V \hat{f}_k).
\end{equation}
which is subsequently fed into an MLP classifier to predict news veracity. The prediction is formulated as:
\begin{equation}
    \hat{y} = \sigma(\mathbf{W}_c f_{\text{final}} + b_c).
\end{equation}
where $\mathbf{W}_c$ and $b_c$ are the weight and bias of the classifier. Finally, we utilize the binary cross-entropy loss for optimization:
\begin{equation}
    \mathcal{L}_{\text{final}} = - \frac{1}{N} \sum_{i=1}^{N} \left[ y_i \log(\hat{y}_i) + (1 - y_i) \log(1 - \hat{y}_i) \right].
\end{equation}

\section{Experiments}

\begin{table*}[htbp]
    \centering
    
    \caption{Comparison of DIVER with the latest multi-domain fake news detection methods. The superscripts in the DIVER row indicate the performance improvement over the best baseline.}
    \label{tab:main_results}
    \resizebox{\textwidth}{!}{%
    \begin{tabular}{llrrrrrrrrrrrr} 
        \toprule
        \rowcolor{HeaderColor}
        & & \multicolumn{4}{c}{\textbf{Weibo}} & \multicolumn{4}{c}{\textbf{Weibo21}} & \multicolumn{4}{c}{\textbf{GossipCop}} \\
        \cmidrule(lr){3-6} \cmidrule(lr){7-10} \cmidrule(lr){11-14} 
        \rowcolor{HeaderColor}
        \hspace{2.5em}\textbf{Category} & \textbf{Method} & \textbf{Acc(\%)}\textcolor{red}{$\uparrow$} & \textbf{F1-Fake(\%)}\textcolor{red}{$\uparrow$} & \textbf{F1-Real(\%)}\textcolor{red}{$\uparrow$} & \textbf{AUC(\%)}\textcolor{red}{$\uparrow$} & \textbf{Acc(\%)}\textcolor{red}{$\uparrow$} & \textbf{F1-Fake(\%)}\textcolor{red}{$\uparrow$} & \textbf{F1-Real(\%)}\textcolor{red}{$\uparrow$} & \textbf{AUC(\%)}\textcolor{red}{$\uparrow$} & \textbf{Acc(\%)}\textcolor{red}{$\uparrow$} & \textbf{F1-Fake(\%)}\textcolor{red}{$\uparrow$} & \textbf{F1-Real(\%)}\textcolor{red}{$\uparrow$} & \textbf{AUC(\%)}\textcolor{red}{$\uparrow$} \\
        \midrule
        \rowcolor{DataColor}
        & EANN & 82.7 & 82.9 & 82.5 & 87.3 & 87.0 & 86.2 & 87.5 & 89.4 & 86.4 & 59.4 & 92.0 & 85.2 \\
        \rowcolor{DataColor}
        & FND-CLIP & 90.7 & 90.8 & 90.7 & 95.3 & 94.3 & 94.0 & 94.6 & 96.2 & 88.0 & 63.8 & 92.8 & 87.1 \\
        \rowcolor{DataColor}
        & MIMoE-FND & 92.8 & 92.8 & 92.8  & \second{97.2} & \second{95.6} & \second{95.5} & \second{95.7} & \second{97.7} & 89.5 & 69.8 & \second{93.8} & 87.9 \\
        \rowcolor{DataColor}
        \multirow{-4}{*}{\shortstack{Multimodal \\ multi-domain methods}}
        & KEN & \second{93.5} & 93.5 & 93.4 & 96.7 & 93.5 & 93.7 & 93.2 & 97.1 & 88.1 & 64.6 & 92.8 & 87.3 \\
        \midrule
        \rowcolor{DataColor}
        & RaCMC & 91.5 & 91.7 & 91.4 & 92.1 & 94.2 & 93.8 & 94.3 & 96.2 & 87.9 & 64.1 & 92.7 & 83.8 \\
        \rowcolor{DataColor}
        & SAFE & 76.2 & 77.4 & 74.8 & 82.4 & 90.5 & 90.1 & 89.0 & 93.7 & 87.2 & 60.3 & 92.4 & 84.3 \\
        \rowcolor{DataColor}
        & CAFE & 84.0 & 84.2 & 83.7 & 89.2 & 88.2 & 88.5 & 87.6 & 90.9 & 86.7 & 58.7 & 92.1 & 85.2 \\
        \rowcolor{DataColor}
        & BMR & 91.8 & 91.4 & 90.4 & 95.4 & 92.9 & 92.7 & 92.5 & 96.2 & 89.5 & 69.1 & 87.6 & \second{88.1} \\
        \rowcolor{DataColor}
        \multirow{-5}{*}{\shortstack{Multimodal \\ single-domain methods}}
        & SEER & 92.9 & 92.8 & \second{93.9} & 93.4 & 93.2 & 92.7 & 92.5 & 96.0 & 89.3 & 67.3 & 87.1 & 87.5 \\
        \midrule
        \rowcolor{DataColor}
        & INSIDE & 88.1 & 68.4 & 93.2 & 91.0 & 89.6 & 81.6 & 83.2 & 87.1 & \second{90.0} & \second{70.7} & 93.4 & \second{88.1} \\
        \rowcolor{DataColor}
        & GLPN-LLM & 92.0 & 93.9 & \second{93.9} & 95.4 & 92.5 & 93.7 & 92.4 & 95.9 & 89.0 & 68.2 & 93.3 & 86.4 \\
        \rowcolor{DataColor}
        \multirow{-3}{*}{\shortstack{Detection methods \\ that reasoning LLMs}}
        & LIFE & 92.4 & \second{94.0} & 92.1 & 95.8 & 92.8 & 93.5 & 93.4 & 97.5 & 86.4 & 70.1 & 86.6 & 85.5 \\
        \midrule
        \rowcolor{DataColor}
        & Qwen 2.5-7B & 93.1 & 93.3 & 93.0 & 96.5 & 93.8 & 93.9 & 93.7 & 96.8 & 88.5 & 66.5 & 93.0 & 86.8 \\
        \rowcolor{DataColor}
        & Qwen3-8B & 93.6 & 93.8 & 93.4 & 96.9 & 94.2 & 94.0 & 94.3 & 97.0 & 89.0 & 68.0 & 93.2 & 87.0 \\
        \rowcolor{DataColor}
        & Llama3.1-8b & 89.5 & 89.0 & 89.8 & 94.5 & 90.5 & 90.2 & 90.8 & 95.0 & 89.8 & 70.5 & 93.6 & 88.0 \\
        \rowcolor{DataColor}
        \multirow{-4}{*}{\shortstack{Large Language \\ Models}}
        & Llama3.2-3B & 87.2 & 86.5 & 87.8 & 92.5 & 88.4 & 88.0 & 88.7 & 93.2 & 88.2 & 65.4 & 92.5 & 86.5 \\
        \midrule
        \rowcolor{DataColor}
        & DIVER 
        & \textcolor{red}{\best{95.8}} & \textcolor{red}{\best{95.2}} & \textcolor{red}{\best{95.7}} & \textcolor{red}{\best{98.2}} 
        & \textcolor{red}{\best{97.7}} & \textcolor{red}{\best{96.6}} & \textcolor{red}{\best{97.0}} & \textcolor{red}{\best{98.8}} 
        & \textcolor{red}{\best{91.6}} & \textcolor{red}{\best{72.3}} & \textcolor{red}{\best{95.4}} & \textcolor{red}{\best{89.3}} \\
        \rowcolor{DataColor}
        \multicolumn{1}{c}{\multirow{-2}{*}{Ours}} & Improv. 
        & 2.46\%$\uparrow$ & 1.28\%$\uparrow$ & 1.92\%$\uparrow$ & 1.03\%$\uparrow$ 
        & 2.20\%$\uparrow$ & 1.15\%$\uparrow$ & 1.36\%$\uparrow$ & 1.13\%$\uparrow$ 
        & 1.78\%$\uparrow$ & 2.26\%$\uparrow$ & 1.71\%$\uparrow$ & 1.36\%$\uparrow$ \\
        \bottomrule
    \end{tabular}%
    }
\end{table*}

\subsection{Experimental Settings}
\indent\textbf{Datasets.} We use three benchmarks: Weibo \cite{wang2018eann}, Weibo21 \cite{zhou2020safe}, and GossipCop \cite{liu2025modality}, following established protocols. Weibo  includes 7,532 training (3,749 real/3,783 fake) and 1,996 test (996 real/1,000 fake) articles. Weibo21  has 9,127 total articles (4,640 real/4,487 fake). GossipCop provides 10,010 training (7,974 real/2,036 fake) and 2,830 testing (2,285 real/545 fake) instances.

\indent\textbf{Baselines.}
We compare our method against three baseline categories: (1) Unimodal methods (MVAN \cite{ni2021mvan}, SpotFake \cite{singhal2019spotfake}); (2) Cross-domain generalization (EANN \cite{wang2018eann}, FND-CLIP \cite{zhou2023multimodal}, MIMoE-FND \cite{liu2025modality},KEN \cite{zhu2025ken}); (3) LLM reasoning (GLPN-LLM \cite{hu2025synergizing}, INSIDE \cite{wang2025bridging}, LIFE \cite{wang2025prompt}).

\textbf{Metrics.} We employ standard evaluation metrics to assess the performance of all methods, including Accuracy (Acc) and F1-score. Following established protocols, we primarily utilize Accuracy as the principal metric for the GossipCop. For the Weibo and Weibo21, we focus on the F1-score (specifically F1-Real and F1-Fake) to evaluate the model's effectiveness. Additionally, we report the Area Under the Curve (AUC) to provide a comprehensive assessment of detection robustness across all benchmarks. For all these metrics, higher values indicate superior performance.

\indent \textbf{Implementation Details.}
Visual features are extracted using CLIP (ViT-B/32) \cite{radford2021learning}. For textual inputs, we employ \texttt{bert-base-chinese} \cite{devlin2019bert} for Chinese datasets (e.g., Weibo and Weibo21) and \texttt{bert-base-uncased} \cite{devlin2019bert} for GossipCop, with a maximum sequence length of 197 tokens. We fine-tune Llama3-8B as the $\text{LLM}_{\text{analyst}}$ \cite{yang2022chinese} using the training splits of the target datasets, where each news instance is converted into instruction-style supervision for claim extraction, consistency verification, and evidence-aware reasoning. During inference, the LLM operates with a temperature of 0.4 to balance determinism and reasoning flexibility, and a maximum context length of 4096 tokens, which is extended to 8192 tokens when iterative visual evidence is injected. In Section 3.4, we employ four latest models (i.e., “OCR”, “Image
Captioning”, “Dense Captioning”, and “Image Tagging”) from Anthropic API as the image-to-text models.The framework is implemented in PyTorch and trained on a single NVIDIA Tesla A100 GPU for 50 epochs with early stopping. The alignment threshold $\beta$ in Phase~3 is selected via grid search on the validation set. The Self-Correction Mechanism iterates for at most $\tau$ rounds, empirically set to $\tau = 2$.

\subsection{Overall Performance} 
We evaluate our framework against 12 strong baselines on three widely used datasets: Weibo, Weibo21 and GossipCop. The comprehensive experimental results for all methods are presented in Table \ref{tab:main_results}.

From the results, we observe that our DIVER framework consistently outperforms existing SoTA methods on all datasets, demonstrating its effectiveness. Specifically, DIVER achieves accuracies of 94.8\%, 96.2\%, and 91.6\% on Weibo, Weibo21, and GossipCop, respectively, which are 2.46\%, 2.20\%, and 3.50\% higher than the previous best methods (i.e., KEN and MIMoE-FND).

\begin{table*}[t]
    \centering
    \caption{Ablation study results showing performance drops for different model variants. We report Accuracy (Acc) and Area Under the Curve (AUC). Module Abbreviations: Embed. (Embedding Model), Vis.Evid. (Visual Evidence), Feedbk (Feedback Mechanism), Facts} (Factual Claims Extraction).
    \label{tab:ablation_moda_llm}
    
    \resizebox{\textwidth}{!}{%
    \begin{tabular}{lcccccccccccc}
        \toprule
        \multirow{2}{*}{Model} & \multicolumn{4}{c}{\textbf{Weibo}} & \multicolumn{4}{c}{\textbf{Weibo-21}} & \multicolumn{4}{c}{\textbf{GossipCop}} \\
        \cmidrule(lr){2-5} \cmidrule(lr){6-9} \cmidrule(lr){10-13}
        & \textbf{Acc} & \textbf{F1-Fake} & \textbf{F1-Real} & \textbf{AUC} & \textbf{Acc} & \textbf{F1-Fake} & \textbf{F1-Real} & \textbf{AUC} & \textbf{Acc} & \textbf{F1-Fake} & \textbf{F1-Real} & \textbf{AUC} \\
        \midrule
        DIVER (Ours) & 95.8 & 95.2 & 95.7 & 98.2 & 97.7 & 96.6 & 97.0 & 98.8 & 91.6 & 72.3 & 95.4 & 89.3 \\
        \midrule
        \multicolumn{13}{c}{\textit{Linguistic Investigation}} \\
        \midrule
        - w/o \text{Prompts} 
        & 92.4\,\textcolor{red!80!black}{\tiny{$\downarrow$3.4\%}} & 92.3\,\textcolor{red!80!black}{\tiny{$\downarrow$2.9\%}} & 92.6\,\textcolor{red!80!black}{\tiny{$\downarrow$3.1\%}} & 94.2\,\textcolor{red!80!black}{\tiny{$\downarrow$4.0\%}}
        & 93.5\,\textcolor{red!80!black}{\tiny{$\downarrow$4.2\%}} & 93.2\,\textcolor{red!80!black}{\tiny{$\downarrow$3.4\%}} & 93.6\,\textcolor{red!80!black}{\tiny{$\downarrow$3.4\%}} & 93.1\,\textcolor{red!80!black}{\tiny{$\downarrow$5.7\%}}
        & 88.1\,\textcolor{red!80!black}{\tiny{$\downarrow$3.5\%}} & 68.7\,\textcolor{red!80!black}{\tiny{$\downarrow$3.6\%}} & 91.2\,\textcolor{red!80!black}{\tiny{$\downarrow$4.2\%}} & 85.8\,\textcolor{red!80!black}{\tiny{$\downarrow$3.5\%}} \\
        - w/o $\text{LLM}_{\text{analyst}}$ 
        & 92.5\,\textcolor{red!80!black}{\tiny{$\downarrow$3.3\%}} & 92.2\,\textcolor{red!80!black}{\tiny{$\downarrow$3.0\%}} & 92.9\,\textcolor{red!80!black}{\tiny{$\downarrow$2.8\%}} & 93.8\,\textcolor{red!80!black}{\tiny{$\downarrow$4.4\%}}
        & 93.1\,\textcolor{red!80!black}{\tiny{$\downarrow$4.6\%}} & 93.4\,\textcolor{red!80!black}{\tiny{$\downarrow$3.2\%}} & 94.0\,\textcolor{red!80!black}{\tiny{$\downarrow$3.0\%}} & 92.6\,\textcolor{red!80!black}{\tiny{$\downarrow$6.2\%}}
        & 87.8\,\textcolor{red!80!black}{\tiny{$\downarrow$3.8\%}} & 67.9\,\textcolor{red!80!black}{\tiny{$\downarrow$4.4\%}} & 92.4\,\textcolor{red!80!black}{\tiny{$\downarrow$3.0\%}} & 85.3\,\textcolor{red!80!black}{\tiny{$\downarrow$4.0\%}} \\
        \midrule
        \multicolumn{13}{c}{\textit{Intra-modal Consistency Reflection}} \\
        \midrule
        - w/o \text{Embed.} 
        & 92.9\,\textcolor{red!80!black}{\tiny{$\downarrow$2.9\%}} & 93.1\,\textcolor{red!80!black}{\tiny{$\downarrow$2.1\%}} & 92.6\,\textcolor{red!80!black}{\tiny{$\downarrow$3.1\%}} & 94.6\,\textcolor{red!80!black}{\tiny{$\downarrow$3.6\%}}
        & 94.5\,\textcolor{red!80!black}{\tiny{$\downarrow$3.2\%}} & 93.8\,\textcolor{red!80!black}{\tiny{$\downarrow$2.8\%}} & 94.1\,\textcolor{red!80!black}{\tiny{$\downarrow$2.9\%}} & 93.3\,\textcolor{red!80!black}{\tiny{$\downarrow$5.5\%}}
        & 88.4\,\textcolor{red!80!black}{\tiny{$\downarrow$3.2\%}} & 69.1\,\textcolor{red!80!black}{\tiny{$\downarrow$3.2\%}} & 90.8\,\textcolor{red!80!black}{\tiny{$\downarrow$4.6\%}} & 85.9\,\textcolor{red!80!black}{\tiny{$\downarrow$3.4\%}} \\
        - w/o \text{Judge} 
        & 93.0\,\textcolor{red!80!black}{\tiny{$\downarrow$2.8\%}} & 92.5\,\textcolor{red!80!black}{\tiny{$\downarrow$2.7\%}} & 92.4\,\textcolor{red!80!black}{\tiny{$\downarrow$3.3\%}} & 94.2\,\textcolor{red!80!black}{\tiny{$\downarrow$4.0\%}}
        & 94.3\,\textcolor{red!80!black}{\tiny{$\downarrow$3.4\%}} & 93.5\,\textcolor{red!80!black}{\tiny{$\downarrow$3.1\%}} & 93.3\,\textcolor{red!80!black}{\tiny{$\downarrow$3.7\%}} & 92.8\,\textcolor{red!80!black}{\tiny{$\downarrow$6.0\%}}
        & 88.0\,\textcolor{red!80!black}{\tiny{$\downarrow$3.6\%}} & 67.7\,\textcolor{red!80!black}{\tiny{$\downarrow$4.6\%}} & 92.2\,\textcolor{red!80!black}{\tiny{$\downarrow$3.2\%}} & 85.4\,\textcolor{red!80!black}{\tiny{$\downarrow$3.9\%}} \\
        \midrule
        \multicolumn{13}{c}{\textit{Inter-modal Alignment Verification}} \\
        \midrule
        - w/o $\text{LLM}_{\text{analyst}}$ 
        & 91.2\,\textcolor{red!80!black}{\tiny{$\downarrow$4.6\%}} & 91.9\,\textcolor{red!80!black}{\tiny{$\downarrow$3.3\%}} & 92.2\,\textcolor{red!80!black}{\tiny{$\downarrow$3.5\%}} & 93.5\,\textcolor{red!80!black}{\tiny{$\downarrow$4.7\%}}
        & 91.0\,\textcolor{red!80!black}{\tiny{$\downarrow$6.7\%}} & 92.5\,\textcolor{red!80!black}{\tiny{$\downarrow$4.1\%}} & 92.1\,\textcolor{red!80!black}{\tiny{$\downarrow$4.9\%}} & 92.7\,\textcolor{red!80!black}{\tiny{$\downarrow$6.1\%}}
        & 86.8\,\textcolor{red!80!black}{\tiny{$\downarrow$4.8\%}} & 68.3\,\textcolor{red!80!black}{\tiny{$\downarrow$4.0\%}} & 90.1\,\textcolor{red!80!black}{\tiny{$\downarrow$5.3\%}} & 85.2\,\textcolor{red!80!black}{\tiny{$\downarrow$4.1\%}} \\
        - w/o \text{Clip} 
        & 90.4\,\textcolor{red!80!black}{\tiny{$\downarrow$5.4\%}} & 91.6\,\textcolor{red!80!black}{\tiny{$\downarrow$3.6\%}} & 91.7\,\textcolor{red!80!black}{\tiny{$\downarrow$4.0\%}} & 94.4\,\textcolor{red!80!black}{\tiny{$\downarrow$3.8\%}}
        & 91.7\,\textcolor{red!80!black}{\tiny{$\downarrow$6.0\%}} & 92.5\,\textcolor{red!80!black}{\tiny{$\downarrow$4.1\%}} & 93.4\,\textcolor{red!80!black}{\tiny{$\downarrow$3.6\%}} & 93.3\,\textcolor{red!80!black}{\tiny{$\downarrow$5.5\%}}
        & 87.5\,\textcolor{red!80!black}{\tiny{$\downarrow$4.1\%}} & 69.2\,\textcolor{red!80!black}{\tiny{$\downarrow$3.1\%}} & 90.9\,\textcolor{red!80!black}{\tiny{$\downarrow$4.5\%}} & 86.2\,\textcolor{red!80!black}{\tiny{$\downarrow$3.1\%}} \\
        \midrule
        \multicolumn{13}{c}{\textit{Evidence-driven Visual Forensics}} \\
        \midrule
        - w/o \text{Facts} 
        & 90.8\,\textcolor{red!80!black}{\tiny{$\downarrow$5.0\%}} & 91.7\,\textcolor{red!80!black}{\tiny{$\downarrow$3.5\%}} & 92.4\,\textcolor{red!80!black}{\tiny{$\downarrow$3.3\%}} & 93.6\,\textcolor{red!80!black}{\tiny{$\downarrow$4.6\%}}
        & 92.1\,\textcolor{red!80!black}{\tiny{$\downarrow$5.6\%}} & 92.6\,\textcolor{red!80!black}{\tiny{$\downarrow$4.0\%}} & 93.3\,\textcolor{red!80!black}{\tiny{$\downarrow$3.7\%}} & 92.5\,\textcolor{red!80!black}{\tiny{$\downarrow$6.3\%}}
        & 87.4\,\textcolor{red!80!black}{\tiny{$\downarrow$4.2\%}} & 68.8\,\textcolor{red!80!black}{\tiny{$\downarrow$3.5\%}} & 90.5\,\textcolor{red!80!black}{\tiny{$\downarrow$4.9\%}} & 85.1\,\textcolor{red!80!black}{\tiny{$\downarrow$4.2\%}} \\
        - w/o \text{Vis.Evid.} 
        & 91.6\,\textcolor{red!80!black}{\tiny{$\downarrow$4.2\%}} & 91.4\,\textcolor{red!80!black}{\tiny{$\downarrow$3.8\%}} & 92.2\,\textcolor{red!80!black}{\tiny{$\downarrow$3.5\%}} & 93.1\,\textcolor{red!80!black}{\tiny{$\downarrow$5.1\%}}
        & 93.1\,\textcolor{red!80!black}{\tiny{$\downarrow$4.6\%}} & 92.3\,\textcolor{red!80!black}{\tiny{$\downarrow$4.3\%}} & 92.9\,\textcolor{red!80!black}{\tiny{$\downarrow$4.1\%}} & 92.2\,\textcolor{red!80!black}{\tiny{$\downarrow$6.6\%}}
        & 87.2\,\textcolor{red!80!black}{\tiny{$\downarrow$4.4\%}} & 68.6\,\textcolor{red!80!black}{\tiny{$\downarrow$3.7\%}} & 90.8\,\textcolor{red!80!black}{\tiny{$\downarrow$4.6\%}} & 85.0\,\textcolor{red!80!black}{\tiny{$\downarrow$4.3\%}} \\
        - w/o \text{Feedbk} 
        & 90.7\,\textcolor{red!80!black}{\tiny{$\downarrow$5.1\%}} & 90.9\,\textcolor{red!80!black}{\tiny{$\downarrow$4.3\%}} & 91.8\,\textcolor{red!80!black}{\tiny{$\downarrow$3.9\%}} & 93.7\,\textcolor{red!80!black}{\tiny{$\downarrow$4.5\%}}
        & 92.6\,\textcolor{red!80!black}{\tiny{$\downarrow$5.1\%}} & 92.0\,\textcolor{red!80!black}{\tiny{$\downarrow$4.6\%}} & 92.5\,\textcolor{red!80!black}{\tiny{$\downarrow$4.5\%}} & 92.4\,\textcolor{red!80!black}{\tiny{$\downarrow$6.4\%}}
        & 86.9\,\textcolor{red!80!black}{\tiny{$\downarrow$4.7\%}} & 69.3\,\textcolor{red!80!black}{\tiny{$\downarrow$3.0\%}} & 91.3\,\textcolor{red!80!black}{\tiny{$\downarrow$4.1\%}} & 85.6\,\textcolor{red!80!black}{\tiny{$\downarrow$3.7\%}} \\
        \bottomrule
    \end{tabular}%
    }
\end{table*}

We further analyze the results from three perspectives:

\begin{itemize}
    \item \textbf{Comparison with Multimodal Baselines:} DIVER establishes a new state-of-the-art in detection accuracy, achieving 95.8\%, 97.7\%, and 91.6\% on Weibo, Weibo21, and GossipCop, respectively. This consistent superiority demonstrates that our dynamic gating mechanism effectively mitigates multimodal noise by selectively filtering irrelevant background details, addressing the accuracy bottlenecks caused by indiscriminate fusion in baselines like SAFE.
    \item \textbf{Comparison with LLM-reasoning Methods:} Unlike standard LLM-based approaches (e.g., GLPN-LLM, LIFE) that often struggle with class imbalance, DIVER delivers substantial gains in both F1-Fake (+2.26\% on GossipCop) and F1-Real (+1.71\% on GossipCop). This balanced performance indicates that our iterative visual feedback strategy provides more refined reasoning than single-pass extraction, effectively reducing both false positives and false negatives.
    \item \textbf{Robustness of Reflections:} The robustness of our framework is evidenced by its high AUC scores across all datasets, peaking at 98.2\% on Weibo and 98.8\% on Weibo21. These results highlight the efficacy of our collaborative reflections in ensuring semantic consistency, which minimizes high-confidence hallucinations and enhances the model's capability to reliably distinguish between authentic and fabricated news.
\end{itemize}

\subsection{Ablation Analysis}

To assess each component within DIVER, we conducted an ablation study on Weibo, Weibo21, and GossipCop (Table \ref{tab:ablation_moda_llm}).

\indent{Impact of Linguistic Investigation.} Removing specifically designed prompts or the $\text{LLM}_{\text{analyst}}$ degrades performance. Accuracy drops by $2.5\%$--$3.5\%$ without prompts and up to $3.8\%$ without the $\text{LLM}_{\text{analyst}}$. This confirms that instruction tuning and LLM capabilities are pivotal for establishing a strong textual baseline focused on linguistic cues.

\indent{Necessity of Intra-modal Consistency Reflection.} Verifying textual consistency is essential; removing the embedding model or judge network decreases accuracy by $2.0\%$ on Chinese datasets and $3.6\%$ on GossipCop. Without these, the system struggles to capture semantic nuances, allowing irrelevant candidates to propagate, highlighting the module's role in filtering noise and enforcing coherence before multimodal interaction.

\indent{Significance of Inter-modal Alignment Verification.} This module exhibits the most significant impact. Removing $\text{LLM}_{\text{analyst}}$ or CLIP leads to accuracy drops, reaching $5.4\%$ on Weibo21 and $4.8\%$ on GossipCop. This decline demonstrates that isolated modality processing is insufficient; CLIP alignment and LLM reasoning are critical for detecting cross-modal mismatches.

\indent Furthermore, our ablation study substantiates the critical role of the Inter-modal Alignment Verification module in mitigating noise injection. As evidenced in Table \ref{tab:ablation_moda_llm}, removing the CLIP module, which governs gating decisions-results in significant performance degradation (e.g., accuracy drops by approximately 5.4\% on Weibo21). This confirms that without dynamic filtering based on alignment scores, the model remains susceptible to interference from irrelevant visual background details, validating our strategy of selective activation to block noise propagation.

\begin{figure}[htbp]
    \centering
    \begin{subfigure}{0.49\textwidth}
        \includegraphics[width=\linewidth]{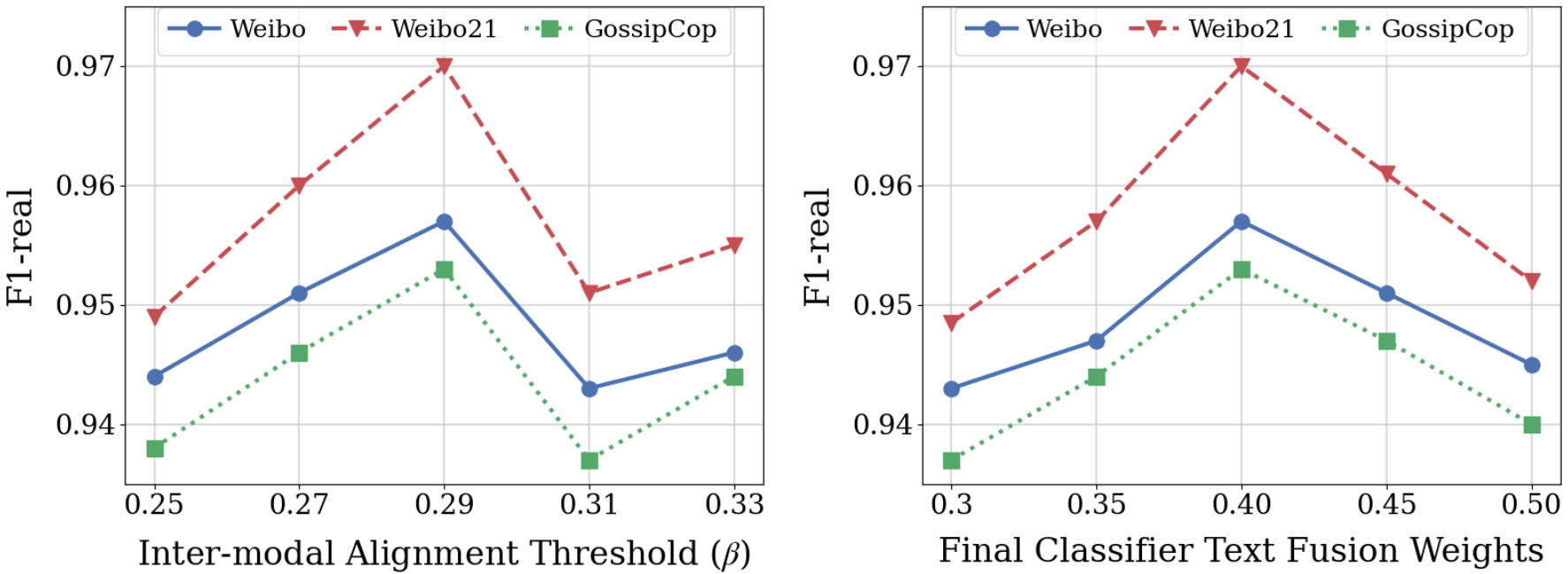}
        \label{fig:hyper_a}
    \end{subfigure}
    \hfill
    \begin{subfigure}{0.49\textwidth}
        \includegraphics[width=\linewidth]{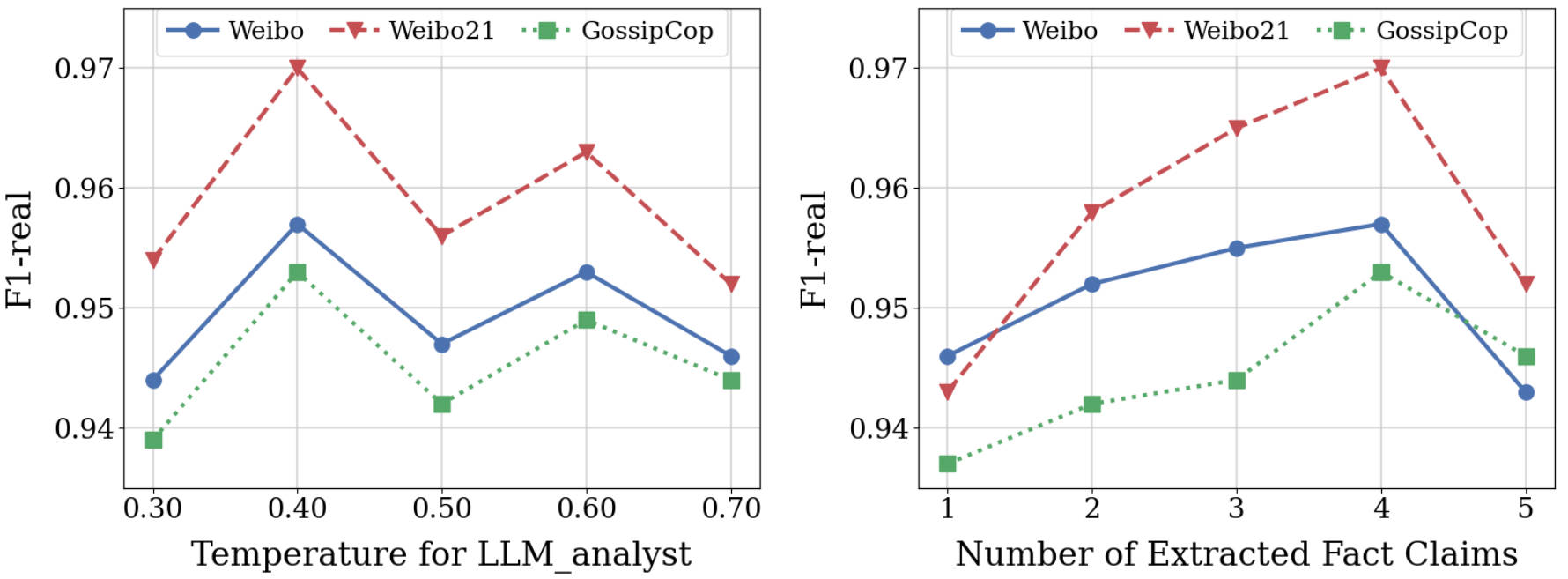}
        \label{fig:hyper_b}
    \end{subfigure}
    \vspace{-1cm}
    \caption{Analysis of hyperparameter sensitivity. This figure shows the impact of four different hyperparameters on the model's F1-real score across three datasets.}
    \label{fig:hyperparameter_sensitivity}
    \vspace{-6pt}
\end{figure}

\section{Discussion}
\subsection{Parameter Sensitivity Analysis}

\textbf{Effect of Evidence-driven Visual Forensics.} Removing factual claims or diverse visual evidence significantly degrades performance, confirming that iteratively introducing specific visual clues is essential for resolving ambiguities when text is insufficient.

We examined sensitivity to four key hyperparameters: the inter-modal alignment threshold ($\beta$), $LLM_{\text{analyst}}$ temperature, text fusion weight, and factual claim count. Figure \ref{fig:hyperparameter_sensitivity} indicates optimal performance at $\beta=0.29$, temperature=0.40, fusion weight=0.40, and 4 claims. Deviations from these values cause only gradual performance declines, demonstrating the framework's robustness. Comprehensive evaluations for F1-Fake, Accuracy, AUC, and the self-correction limit ($\tau=2$) are provided in the Appendix, further confirming consistent stability across metrics.

\subsection{Case Study: Model Explainability}
To qualitatively validate DIVER, we visualize representative samples to show how the model dynamically balances computational efficiency with reasoning depth (full details in Appendix \ref{sec:appendix_case_study}).

Our analysis confirms DIVER adheres to cognitive economy. In scenarios with congruent visual and textual information, the model computes a high alignment score and acts as a fast System 1 verifier, bypassing redundancy. Conversely, in adversarial cases like ambiguous images, it triggers the System 2 Evidence-driven Visual Forensics module. This enables the iterative extraction of fine-grained evidence (e.g., OCR) to explicitly refute contradictions, thereby mitigating LLM hallucinations and ensuring transparent decision-making.

\begin{figure}[htbp]
\centering
\includegraphics[width=0.9\linewidth]{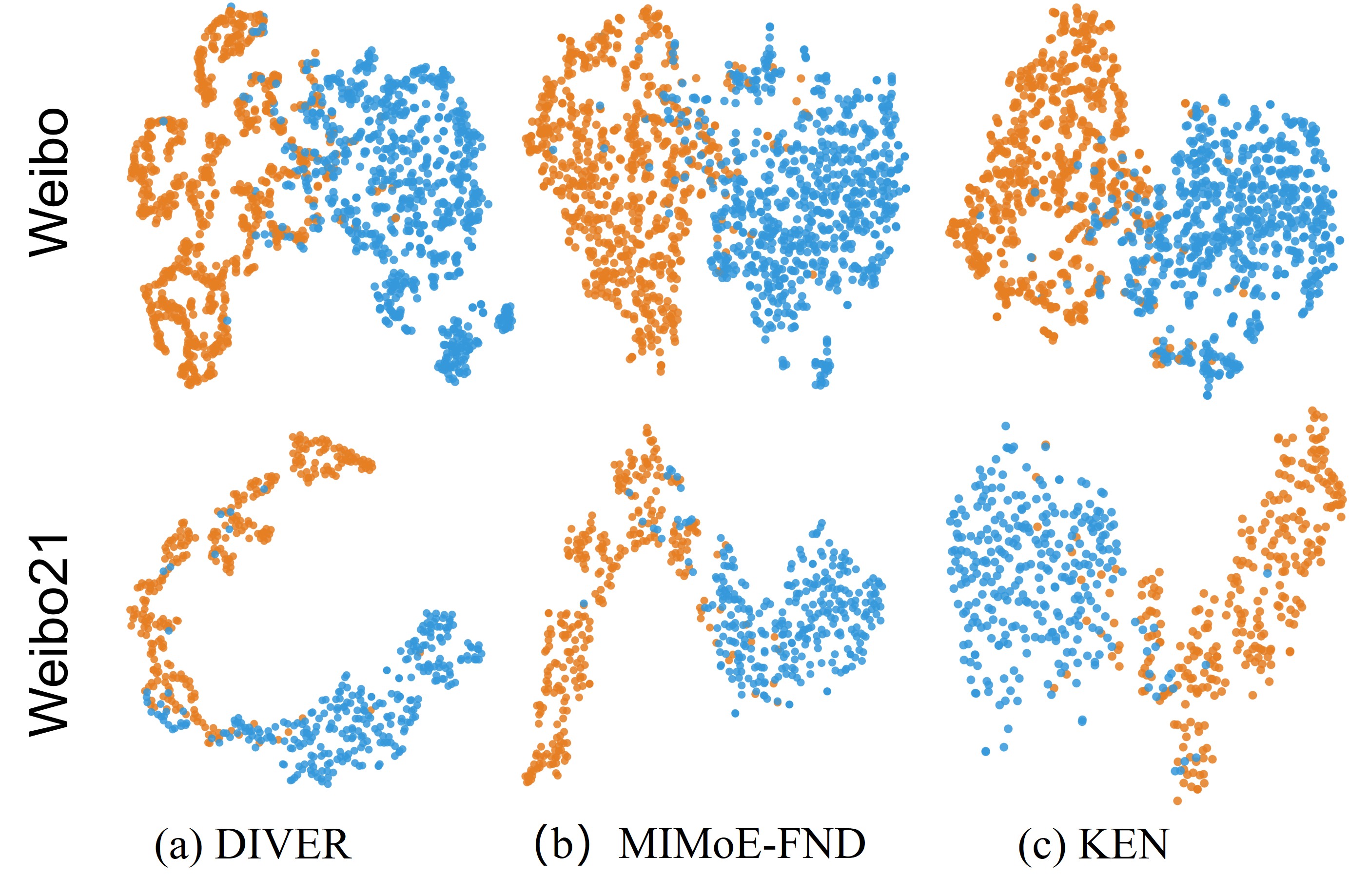}
\caption{T-SNE visualization of test set features. Same color dots indicate the same label.}
\label{fig:tsne_visualization}
\vspace{-6pt}
\end{figure}

\subsection{T-SNE Visualizations}
Figure \ref{fig:tsne_visualization} shows T-SNE visualizations of features from DIVER, MIMoE-FND\cite{liu2025modality}, and KEN\cite{zhu2025ken} on the Weibo and Weibo21 test sets. Compared to the baselines, DIVER produces fewer fake news outliers and less overlap between real and fake news embeddings, confirming its superior performance

\indent On Weibo21, DIVER's features form multiple, separated subclusters, unlike the single clusters on Weibo. This suggests Weibo21 has varying topics and that DIVER not only distinguishes authenticity but also captures deep, event level semantic information, spatially distinguishing events.

\begin{table}[t]
    \centering
    \small
    \caption{Average response time (s) for verifying a single news sample using different LLM-based methods. The values in brackets indicate the accuracy improvement of DIVER compared to the baselines.}
    \label{tab:efficiency}
    \resizebox{\columnwidth}{!}{
        \begin{tabular}{l|ccc|c}
        \toprule
        Model & Weibo & Weibo21 & GossipCop & Avg. \\
        \midrule
        GLPN-LLM & 5.35 & 5.95 & 5.88 & 5.73 \\
        INSIDE & 4.53 & 5.12 & 5.25 & 4.97 \\
        LIFE & 6.21 & 7.45 & 7.60 & 7.09 \\
        \textbf{DIVER (Ours)} & \textbf{3.85} & \textbf{4.20} & \textbf{4.32} & \textbf{4.12} \\
        \bottomrule
        \end{tabular}
    }
    \vspace{-12pt}
\end{table}
\subsection{Inference Efficiency}
To evaluate computational efficiency, we introduce average inference latency per sample (``Avg.''). Table \ref{tab:efficiency} compares DIVER against LLM-based methods including GLPN-LLM, INSIDE, and LIFE.

Observations indicate: (1) DIVER achieves superior performance and the fastest inference speed among comparable methods. It reduces average reasoning time by 2.97 s compared to LIFE, and outperforms INSIDE by 0.80 s. (2) This gain is attributed to the Uncertainty-Aware Dynamic Fusion mechanism. Unlike approaches requiring exhaustive visual forensics, DIVER filters straightforward samples at the Inter-modal Alignment Verification stage. Dynamically bypassing the expensive Stage 4 for samples with high semantic alignment minimizes redundant overhead without compromising reasoning depth. For a detailed cost-benefit analysis and specific computational complexity metrics of this module, please refer to Appendix \ref{sec:appendix_c}.

\section{Conclusion}

This paper proposes a novel framework called DIVER, which leverages intra-modal and inter-modal co-reflection for multimodal fake news detection. DIVER uniquely evaluates text consistency and resolves various conflicts through a visual iterative feedback module. By introducing visual evidence and integrating forensic cues, our method guides large language models to progressively improve cross-modal reasoning. Extensive experiments on Weibo and GossipCop demonstrate that DIVER significantly outperforms state-of-the-art baseline methods, with accurate and easily interpretable validation results.

\section{Limitation.}
We would like to explore the following aspects in limitation: \ding{182} enhancing the framework's generalization to more diverse social media domains, \ding{183} evaluating the framework's scalability in more complex real world scenarios and its robustness against more unforeseen adversarial attacks, and \ding{184} investigating lightweight model distillation techniques to further adapt the framework for resource-constrained edge devices.

\bibliography{custom} 

\newpage

\appendix

\section{Case Study}
\label{sec:appendix_case_study}

In this section, we provide a comprehensive qualitative analysis of the DIVER framework using representative samples from the Weibo dataset. Table \ref{tab:adaptive_structure_transposed_3cases} visualizes the step-by-step inference process, comparing consistent scenarios (Real news) with adversarial scenarios (Fake news). This comparison highlights the model's dynamic mechanism in balancing computational efficiency with forensic depth.

\subsection{Efficiency in Consistent Scenarios}
As illustrated in \textbf{Case 1} and \textbf{Case 2} of Table \ref{tab:adaptive_structure_transposed_3cases}, when the visual and textual modalities are semantically congruent, DIVER effectively conserves computational resources.
\begin{itemize}
    \item \textbf{Case 1 (Real):} The news text describes "Andy Cohen interviews actresses," which aligns well with the generated image caption "Andy Cohen interviews guests on his show." The Inter-modal Alignment Verification module computes a high semantic alignment score ($S_{inter} \ge \beta$), identifying the sample as "safe." Consequently, the model acts as a System 1 verifier and terminates the search early. The deep forensic tools (OCR, Dense Captions) are marked as "Not Reached," avoiding redundant computation.
    \item \textbf{Case 2 (Real):} Similarly, the text regarding "Kate Middleton" matches the visual gist of "Royal family members." The model correctly bypasses the evidence-driven stage, ensuring high inference speed.
\end{itemize}

\begin{table*}[htbp]
  \centering
  \footnotesize
  \vspace{-4pt}

  \caption{Explainability case study. Consistent samples (Cases 1 \& 2) bypass deep forensics (marked "Not Reached") for efficiency. In contrast, the adversarial Case 3 triggers the evidence-driven module, where specific visual clues (e.g., OCR "TEAM USA") are extracted to explicitly refute the contradictory text claim.}

  \label{tab:adaptive_structure_transposed_3cases}
  \renewcommand{\arraystretch}{1.3} 
  \setlength{\tabcolsep}{4pt}        

  \begin{tabular}{c >{\bfseries}l >{\centering\arraybackslash}m{4.5cm} >{\centering\arraybackslash}m{4.5cm} >{\centering\arraybackslash}m{4.5cm}}
  \toprule
    & Keys & \textbf{Case 1 (Real)} & \textbf{Case 2 (Real)} & \textbf{Case 3 (Fake)} \\
  \midrule

  \multirow{5}{*}{\textbf{News}} & \makecell[l]{Image\\News} & 
  \includegraphics[width=2.5cm, height=1.8cm, keepaspectratio]{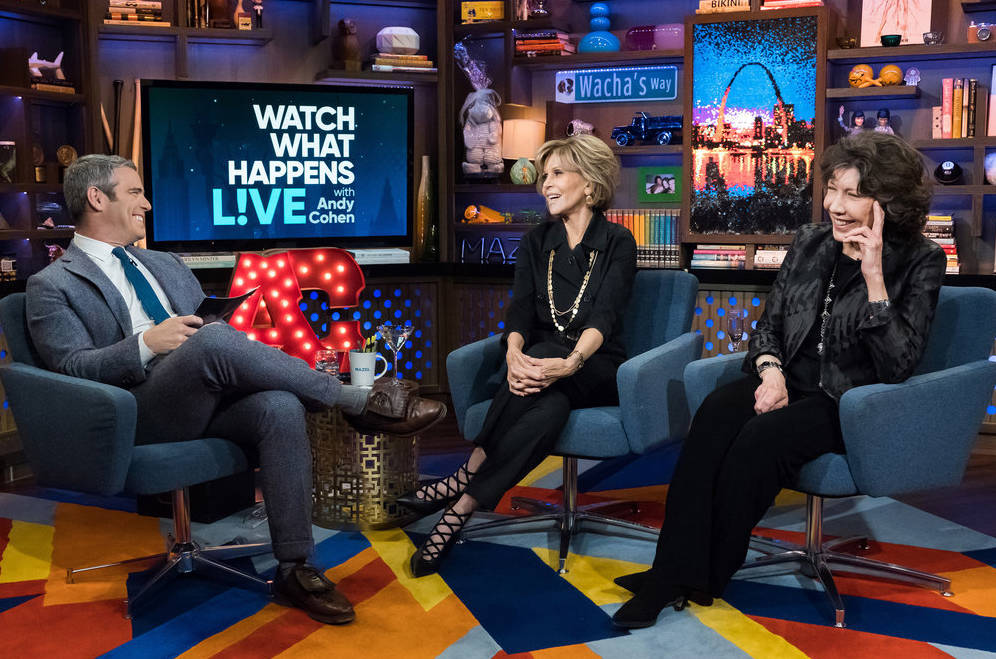} & 
  \includegraphics[width=8.5cm, height=1.8cm, keepaspectratio]{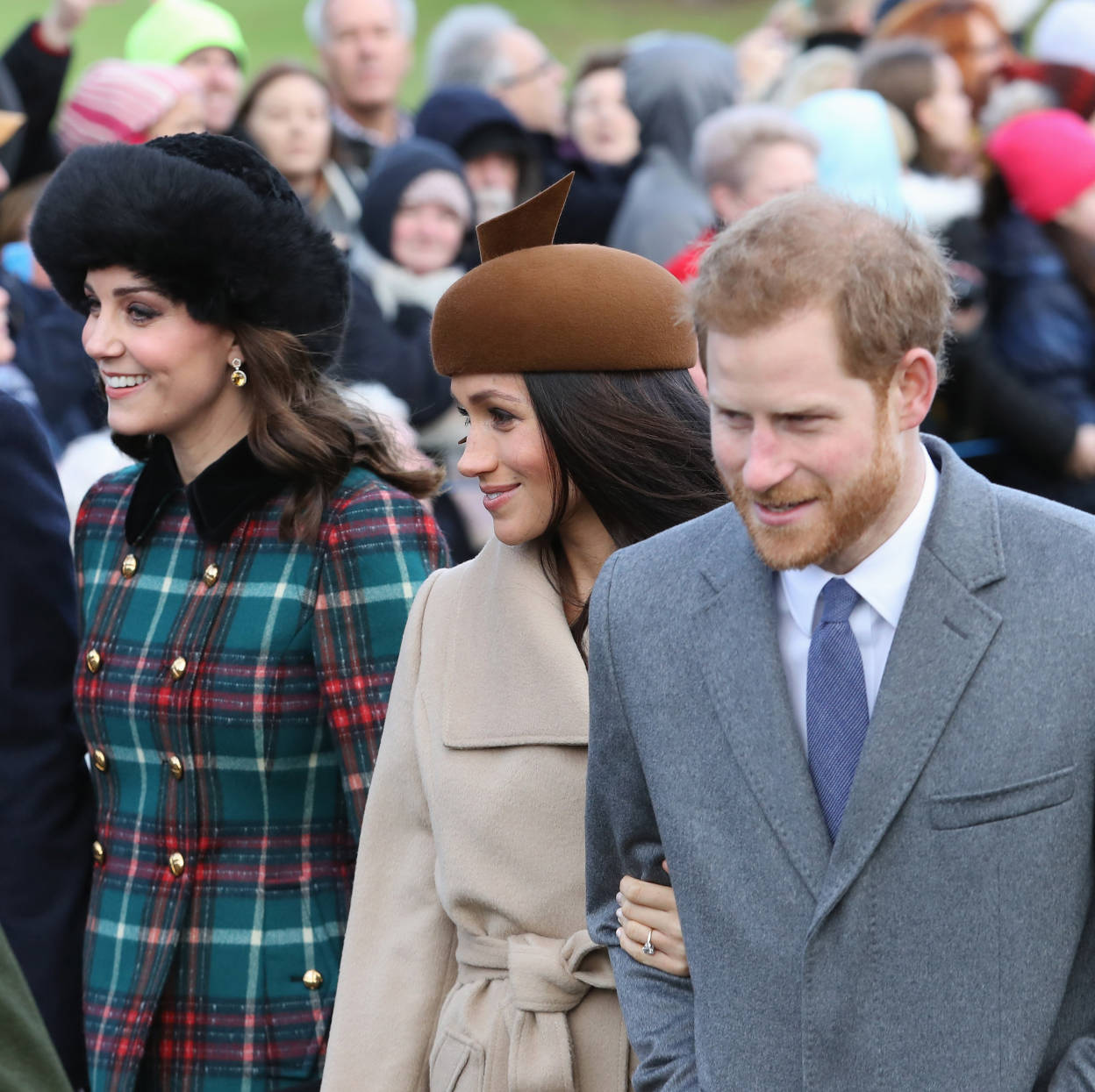} & 
  \includegraphics[width=2.5cm, height=1.8cm, keepaspectratio]{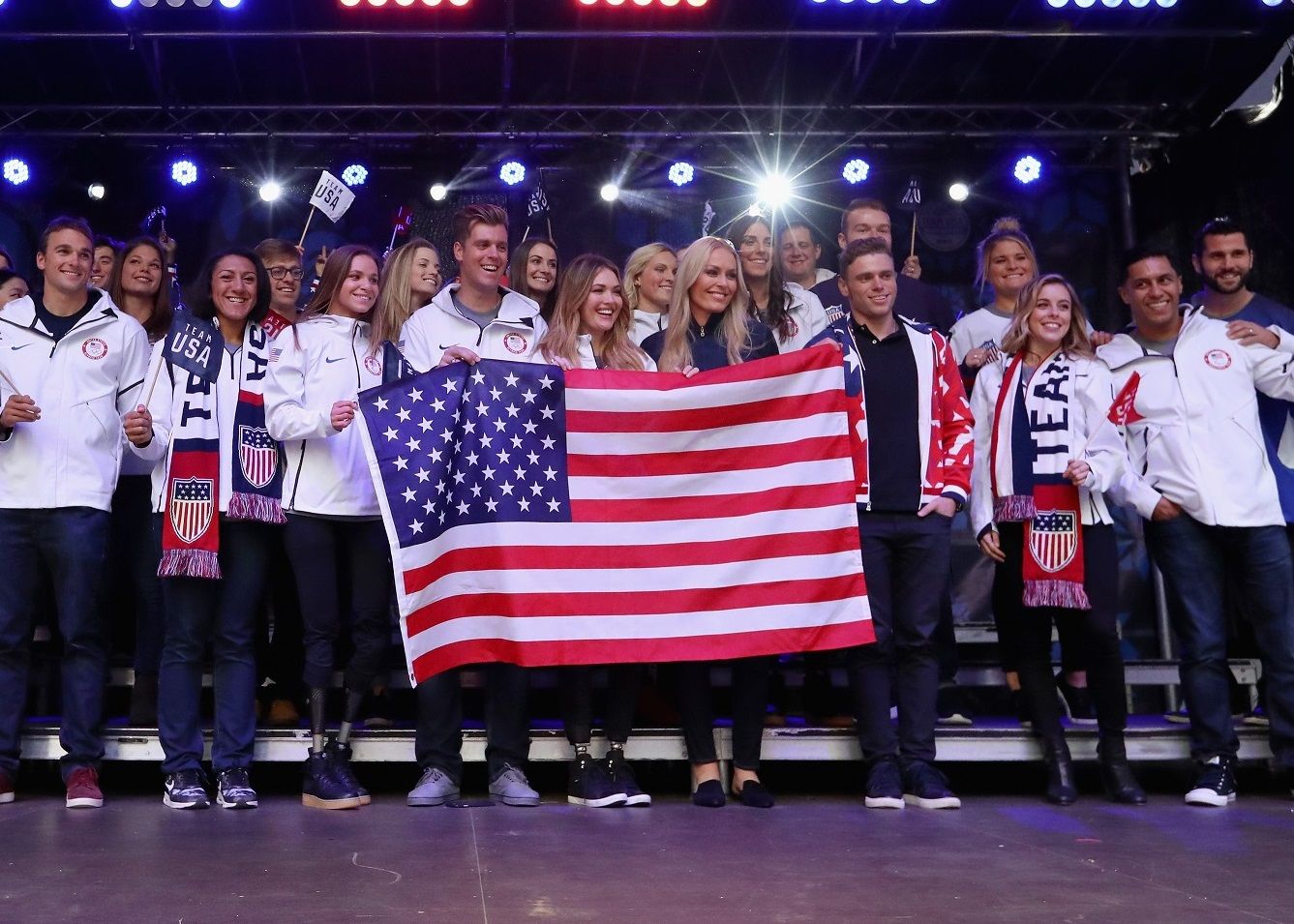} \\

  \cmidrule{2-5}

   & \makecell[c]{Text\\News} & 
  \makecell{Host Andy Cohen interviews \\actresses Jane Fonda\\ and Lily Tomlin} & 
  \makecell{Kate Middleton joins \\ Meghan Markle for \\Christmas service} & 
  \makecell{Brad Pitt attends the annual \\Golden Globe Awards} \\
  \midrule

  \multirow{8}{*}{\textbf{Method}} & \makecell[l]{OCR} & 
  \makecell{Not Reached} & 
  \makecell{Not Reached} & 
  \makecell{TEAM USA \\ (on scarves)} \\
  \cmidrule{2-5}

   & \makecell[l]{Caption} & 
  \makecell{Andy Cohen interviews \\guests on his show} & 
  \makecell{Royal family members \\walk side by side} & 
  \makecell{Not Reached} \\
  \cmidrule{2-5}
   & \makecell[l]{Dense\\Captions} & 
  \makecell{Not Reached} & 
  \makecell{Two women wearing \\hats and coats walking} & 
  \makecell{Smiling people wearing \\ white sports jackets} \\
  \cmidrule{2-5}
   & \makecell[l]{Tags} & 
  \makecell{Talk show, Interview, \\ TV Host} & 
  \makecell{Not Reached} & 
  \makecell{Sports, Olympics, \\ Team USA} \\
  \bottomrule
  \end{tabular}
\vspace{-4pt}
\end{table*}

\subsection{Robustness in Adversarial Scenarios}

In contrast, \textbf{Case 3 (Fake)} demonstrates the necessity of the System 2 reasoning module when dealing with sophisticated cheapfakes.
\begin{itemize}
    \item \textbf{Case 3 (Fake):} The text claims "Brad Pitt attends the annual Golden Globe Awards," while the image depicts a person who superficially resembles him but is actually an athlete. Static fusion models might be misled by the visual similarity. However, DIVER detects a potential mismatch (low semantic alignment score) and triggers the Evidence-driven Visual Forensics module.

\section{Appendix: Additional Sensitivity Analysis}
\label{sec:appendix_metric_sensitivity}
    \item \textbf{Forensic Reasoning:} The model iteratively invokes tools to extract fine-grained evidence. Crucially, the OCR tool detects the text "TEAM USA" on the scarves, and the Image Tags tool identifies "Olympics." These specific visual clues explicitly refute the "Golden Globe Awards" claim.
    \item \textbf{Outcome:} By grounding the decision in this extracted evidence, DIVER prevents LLM hallucination and correctly classifies the news as fake, providing transparent and factual support for its decision.
\end{itemize}

While the main text analyzes hyperparameter sensitivity primarily through \textbf{F1-real}, we further investigate the robustness of our framework from a centric metric perspective. Specifically, we analyze the behavior of \textbf{F1-Fake}, \textbf{Accuracy (Acc)}, and \textbf{Area Under the Curve (AUC)} under variations of the same hyperparameter settings shown in Figure~3.

\begin{figure}[htbp]
    \centering
    \begin{subfigure}{0.49\textwidth}
        \includegraphics[width=\linewidth]{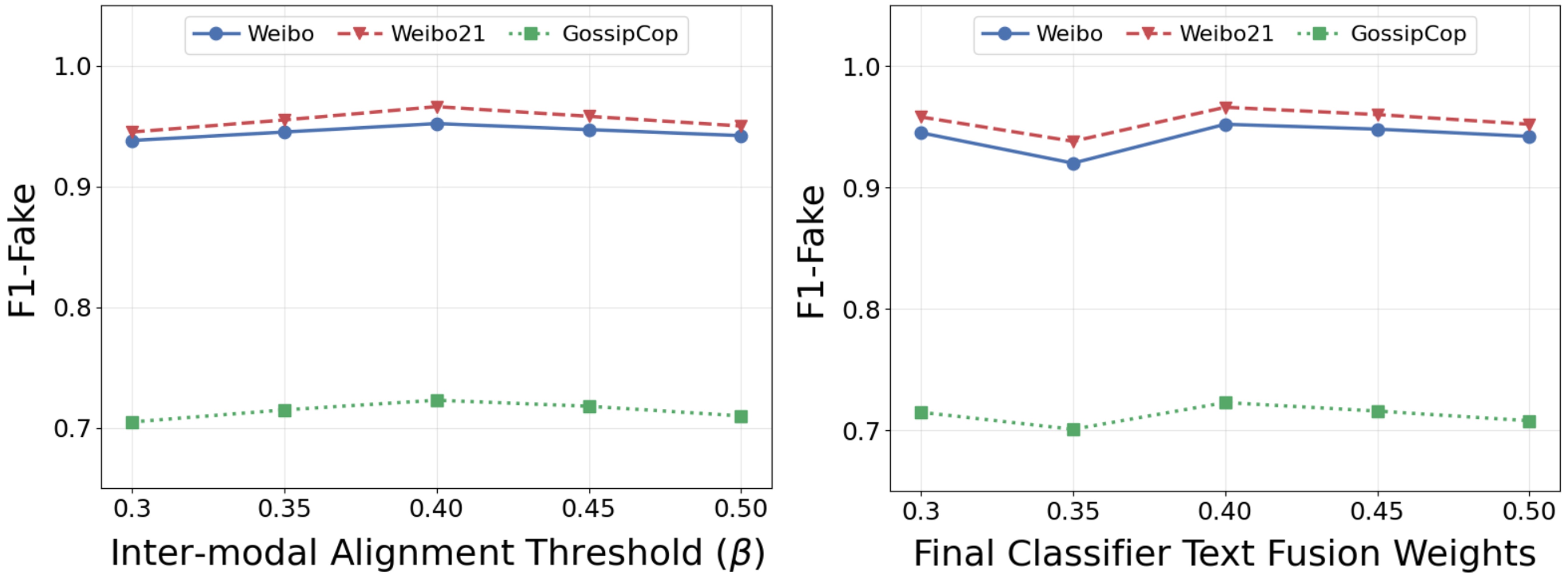}
        \label{fig:hyper_fake_a}
    \end{subfigure}
    \hfill
    \begin{subfigure}{0.49\textwidth}
        \includegraphics[width=\linewidth]{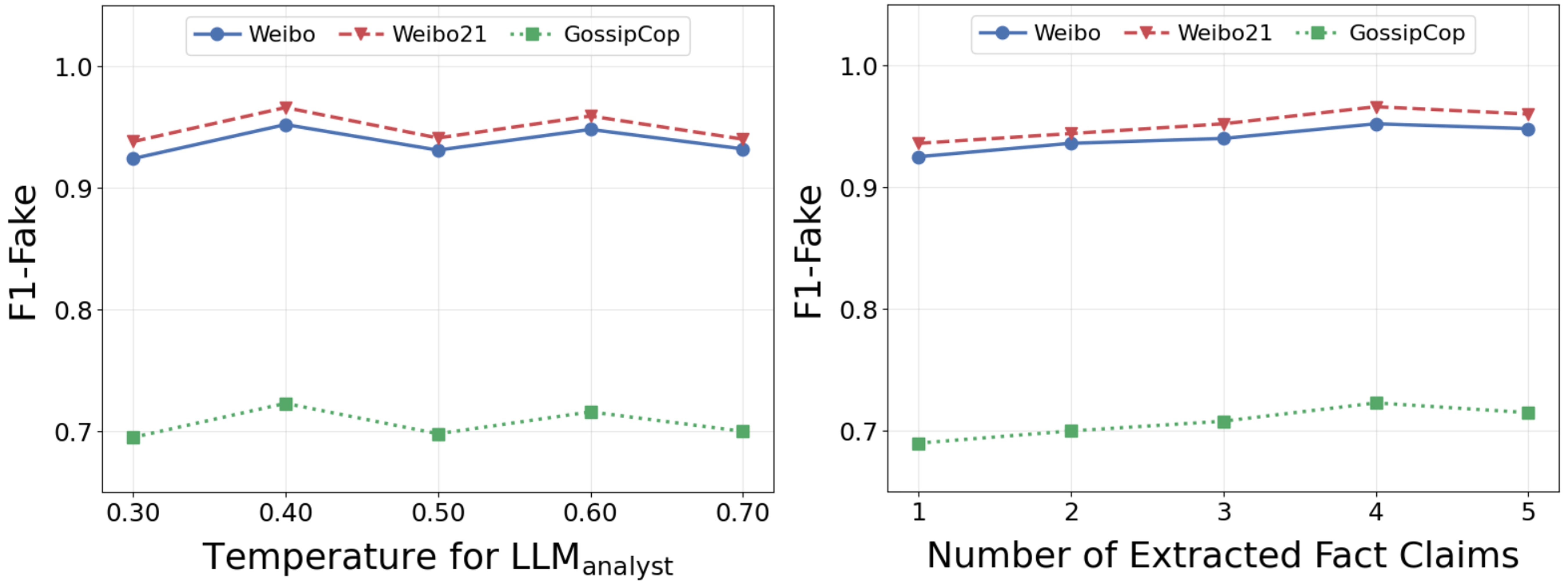}
        \label{fig:hyper_fake_b}
    \end{subfigure}
    \vspace{-1cm}
    \caption{Analysis of hyperparameter sensitivity. This figure illustrates the impact of four different hyperparameters on the model's F1-Fake score across three datasets.}
    \label{fig:hyperparameter_sensitivity_fake}
    \vspace{-6pt}
\end{figure}

\indent \textbf{{Sensitivity of F1-Fake.}}
Across all hyperparameter configurations, F1-Fake exhibits trends that are largely consistent with F1-real, while showing slightly higher sensitivity to parameter deviations. This behavior is expected, as correctly identifying fake news relies more heavily on fine-grained cross-modal inconsistencies and evidence-driven visual forensics. When hyperparameters deviate from their optimal ranges, the decline in F1-Fake is more noticeable than that of F1-real, indicating increased false negatives for adversarial samples. Nevertheless, the performance degradation remains gradual rather than abrupt, demonstrating that the proposed framework maintains stable fake news detection capability even under non-optimal settings.

\indent \textbf{{Sensitivity of Accuracy (Acc).}}
Accuracy remains comparatively stable across a wide range of hyperparameter values. Unlike specific class metrics, Acc reflects the model’s overall decision consistency and is therefore less sensitive to localized performance fluctuations on either real or fake samples. 

\begin{figure}[htbp]
    \centering
    \begin{subfigure}{0.49\textwidth}
        \includegraphics[width=\linewidth]{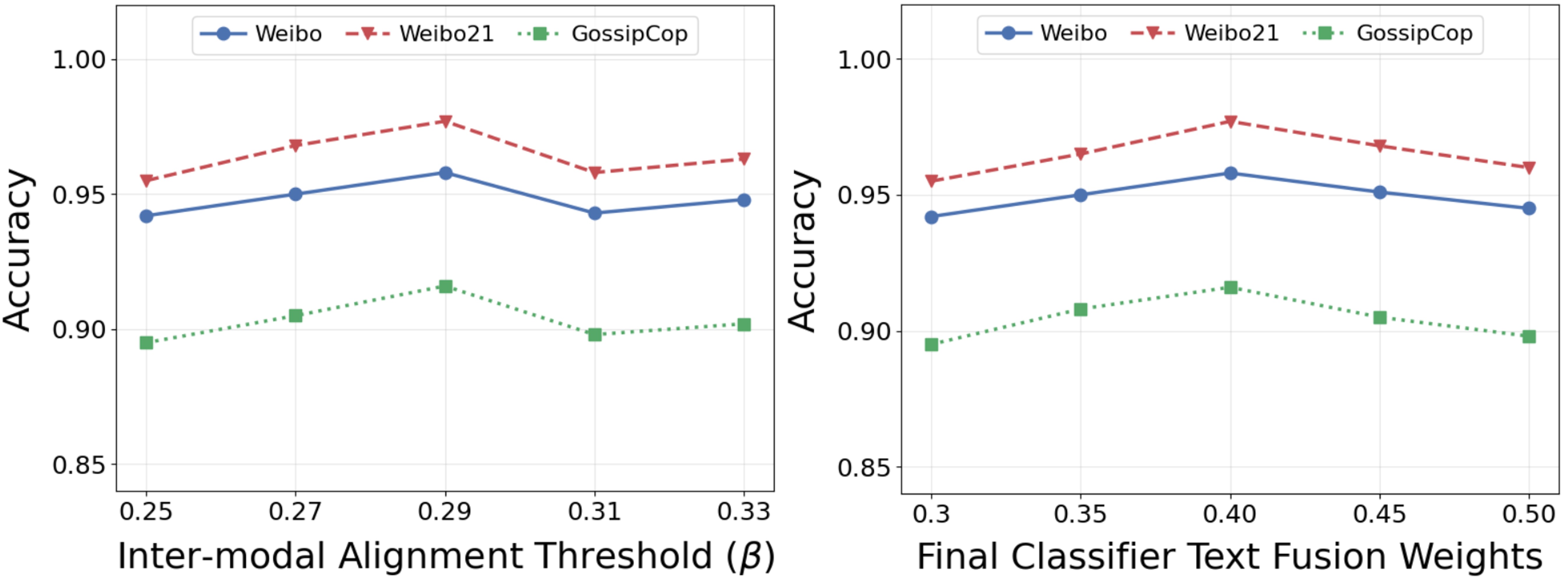}
        \label{fig:hyper_acc_a}
    \end{subfigure}
    \hfill
    \begin{subfigure}{0.49\textwidth}
        \includegraphics[width=\linewidth]{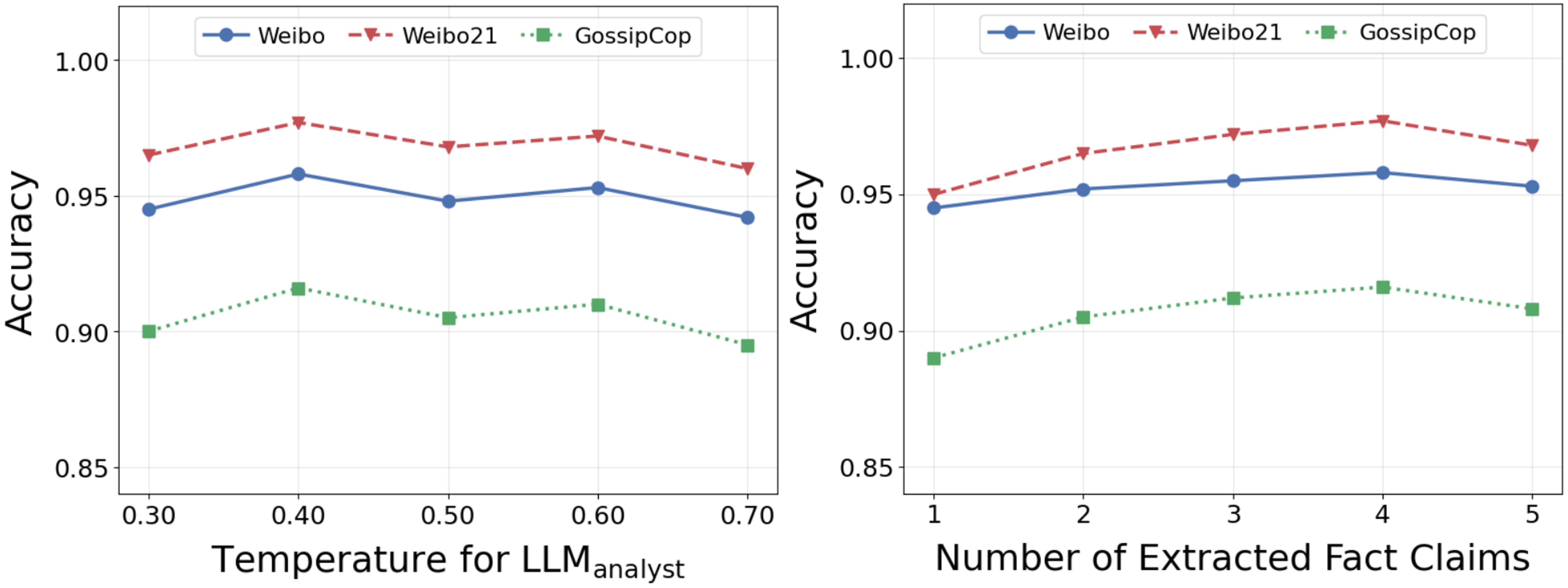}
        \label{fig:hyper_acc_b}
    \end{subfigure}
    \vspace{-1cm}
    \caption{Analysis of hyperparameter sensitivity. This figure illustrates the impact of four different hyperparameters on the model's Accuracy (Acc) across three datasets.}
    \label{fig:hyperparameter_sensitivity_acc}
    \vspace{-6pt}
\end{figure}

The consistently high Acc observed in the sensitivity plots suggests that DIVER preserves reliable global decision boundaries, benefiting from its dynamic gating mechanism that selectively invokes visual forensics only when necessary. This stability indicates that moderate hyperparameter perturbations do not significantly affect the model’s overall classification correctness.

\begin{figure}[htbp]
    \centering
    \begin{subfigure}{0.49\textwidth}
        \includegraphics[width=\linewidth]{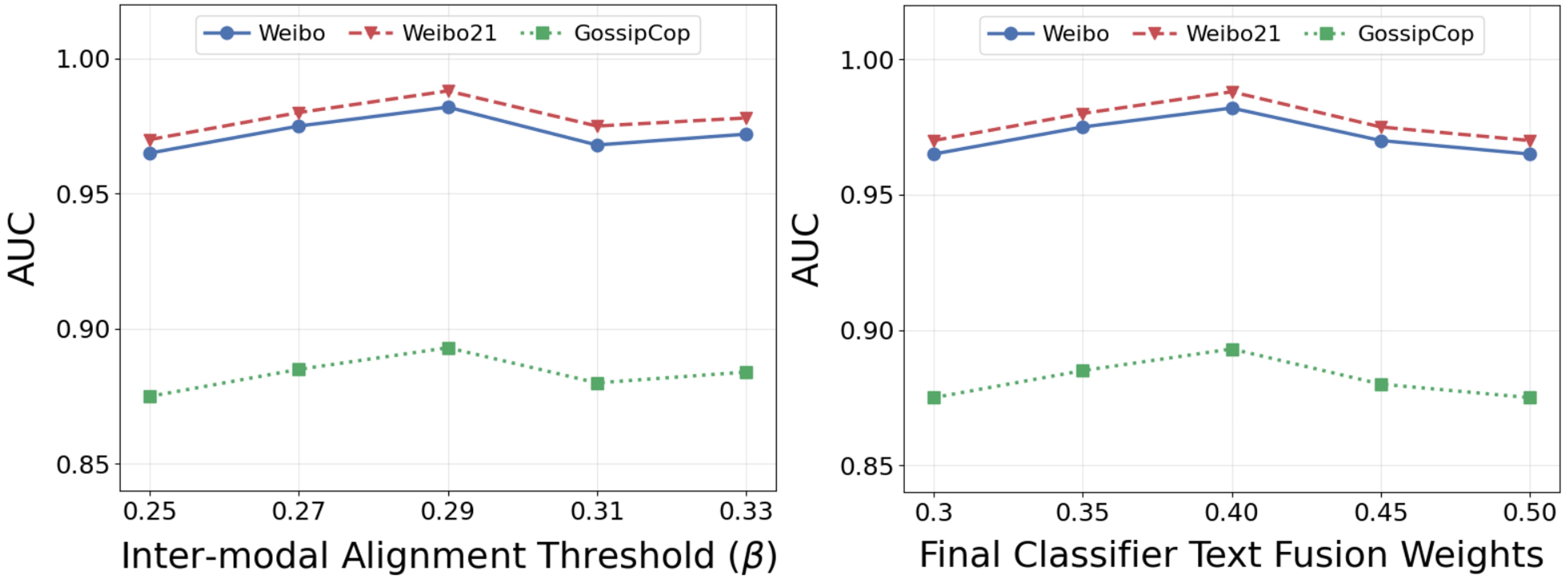}
        \label{fig:hyper_auc_a}
    \end{subfigure}
    \hfill
    \begin{subfigure}{0.49\textwidth}
        \includegraphics[width=\linewidth]{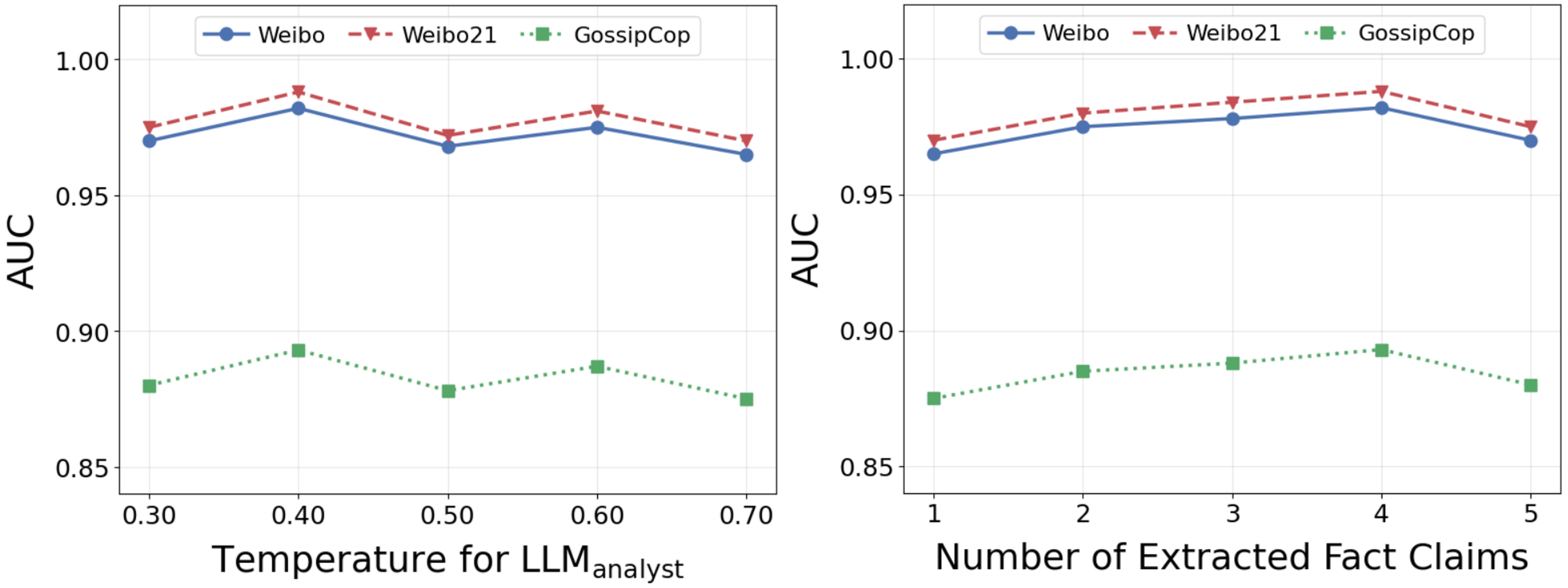}
        \label{fig:hyper_auc_b}
    \end{subfigure}
    \vspace{-1cm}
    \caption{Analysis of hyperparameter sensitivity. This figure illustrates the impact of four different hyperparameters on the model's Area Under the Curve (AUC) across three datasets.}
    \label{fig:hyperparameter_sensitivity_auc}
    \vspace{-6pt}
\end{figure}

\indent \textbf{Sensitivity of AUC.}
AUC demonstrates the strongest robustness among all evaluation metrics. Across all examined hyperparameter settings, AUC curves remain smooth with minimal variance, indicating that the relative ranking between real and fake samples is largely preserved. This suggests that the confidence calibration of DIVER is resilient to hyperparameter changes, even when absolute classification thresholds become suboptimal. The consistently high AUC values further confirm that the proposed uncertainty-aware fusion and iterative evidence refinement mechanisms enhance the model’s discriminative capacity across diverse operating conditions.

\begin{figure}[htbp]
    \centering
    \begin{subfigure}{1.0\linewidth} 
        \includegraphics[width=\linewidth]{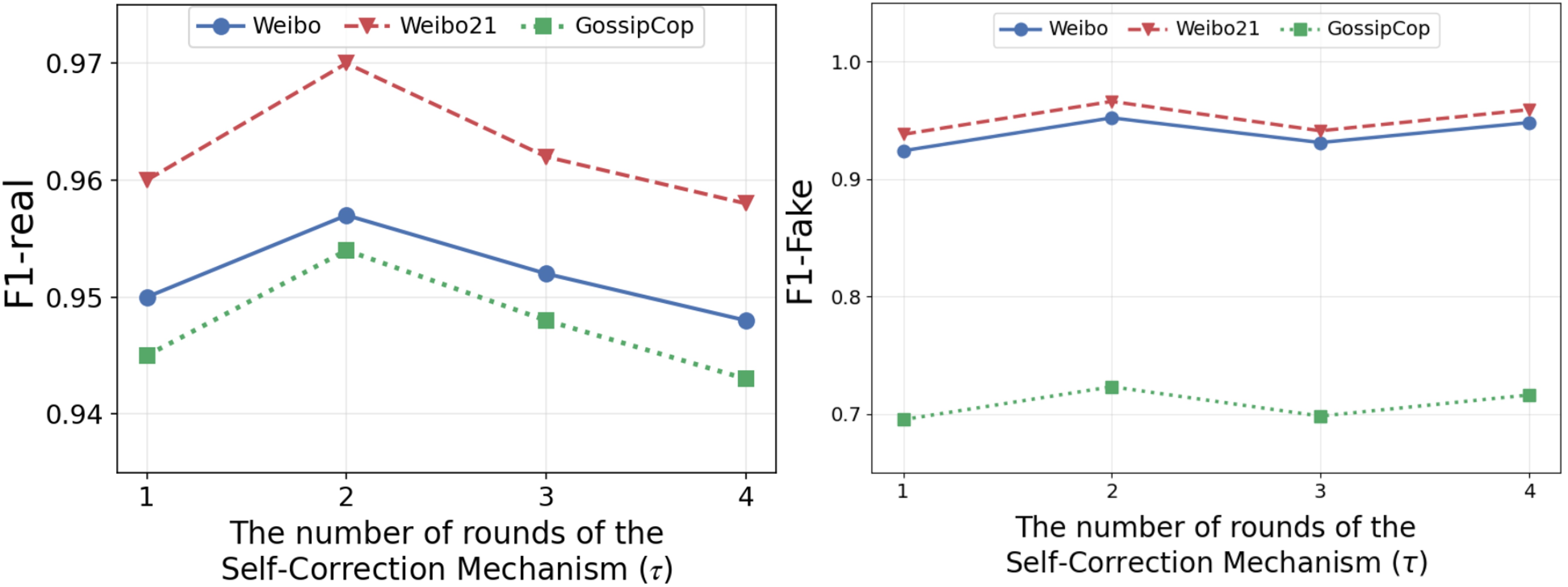}
        \label{fig:tau_f1}
    \end{subfigure}
    
    \vspace{0.2cm} 
    
    \begin{subfigure}{1.0\linewidth}
        \includegraphics[width=\linewidth]{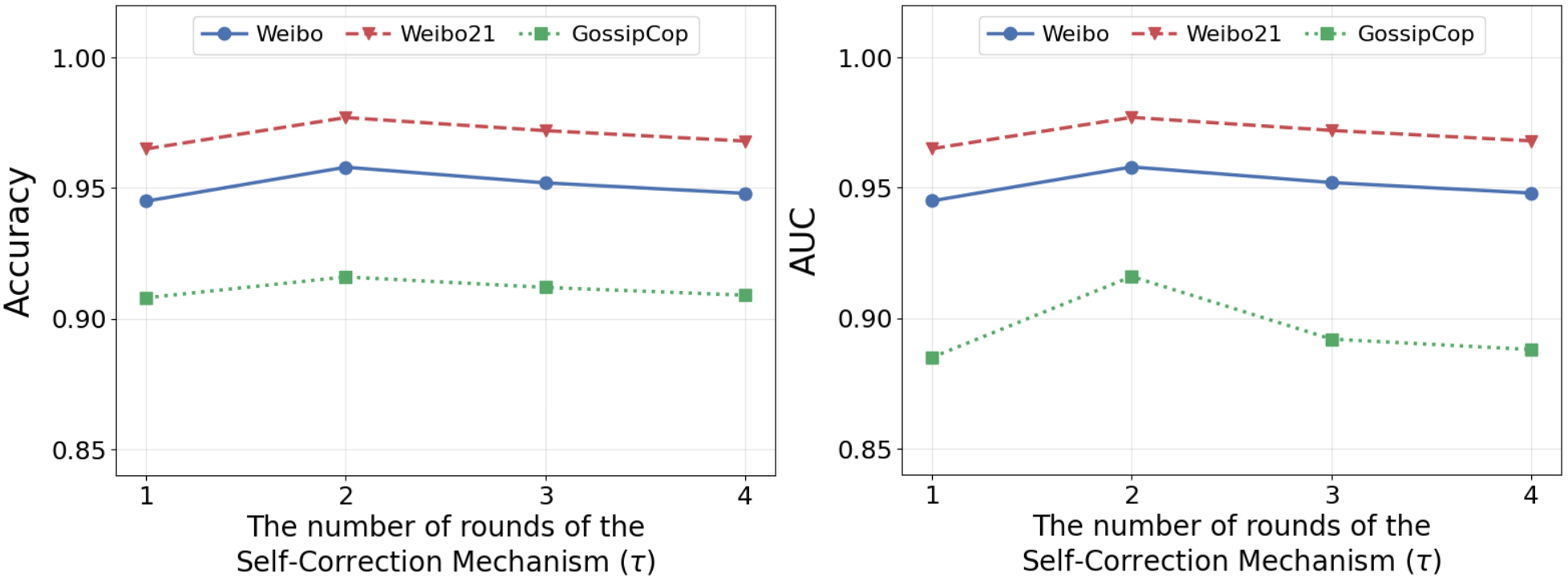}
        \label{fig:tau_acc_auc}
    \end{subfigure}
    
    \vspace{-0.2cm}
    \caption{Sensitivity analysis of the self-correction iteration limit ($\tau$). This figure illustrates the impact of varying correction rounds on F1-scores, Accuracy, and AUC. Performance consistently peaks at $\tau=2$ across all datasets, with degradation observed at $\tau > 2$.}
    \label{fig:tau_sensitivity}
    \vspace{-6pt}
\end{figure}

\indent \textbf{Impact of Self-Correction Iteration Limit ($\tau$).}
To determine the optimal depth for our intra-modal consistency reflection, we investigate the impact of the self-correction iteration limit $\tau$ on model performance. Figure~\ref{fig:tau_sensitivity} illustrates the variations in F1-scores, Accuracy, and AUC across all three datasets as $\tau$ increases from 1 to 4.

\begin{table*}[t]
\centering
\small
\setlength{\tabcolsep}{4.5pt} 
\caption{Cost-Benefit Analysis: Comparison of computational overhead versus performance on Weibo, Weibo21, and GossipCop datasets. While baselines like LIFE \cite{wang2025prompt} and GLPN-LLM \cite{hu2025synergizing} require exhaustive LLM processing for every sample, DIVER employs a dynamic gating mechanism. This allows it to bypass expensive visual forensics for approximately 70\% of samples, resulting in significantly lower amortized costs (API calls and tokens) while achieving state-of-the-art accuracy.}
\label{tab:complexity_analysis}
\begin{tabular}{lccccc}
\toprule
\textbf{Method} & \textbf{Weibo Acc (\%)} & \textbf{Weibo21 Acc (\%)} & \textbf{GossipCop Acc (\%)} & \textbf{Avg. API Calls} & \textbf{Avg. Tokens (k)} \\
\midrule
INSIDE \cite{wang2025bridging} & 88.1 & 93.9 & 87.1 & 1.0 & 1.6 \\
GLPN-LLM \cite{hu2025synergizing} & 92.0 & 94.0 & 88.1 & 1.0 & 1.8 \\
LIFE \cite{wang2025prompt} & 92.4 & 92.1 & 86.6 & 2.0 & 3.2 \\
\midrule
\rowcolor{gray!10} \textbf{DIVER (Ours)} & \textbf{94.8} & \textbf{96.2} & \textbf{91.6} & \textbf{1.3} & \textbf{1.5} \\
\emph{vs. Best Baseline} & \emph{+2.46\%} & \emph{+2.20\%} & \emph{+3.50\%} & \emph{-30\%} & \emph{+6.25\%} \\
\bottomrule
\end{tabular}
\end{table*}

\begin{itemize}
    \item Optimal Convergence at $\tau=2$: We observe a consistent unimodal pattern across all evaluation metrics. The model achieves significant performance gains when transitioning from $\tau=1$ to $\tau=2$, reaching a distinct peak at $\tau=2$. For instance, on the Weibo21 dataset, the F1-real score improves noticeably, indicating that a single round of feedback is often insufficient, while two rounds allow the \textit{Judge Network} to effectively resolve the majority of extractable semantic discrepancies.
    
    \item Performance Degradation with Excessive Correction: Contrary to the assumption that more reasoning yields better results, setting $\tau > 2$ leads to a discernible decline in performance. As shown in the Accuracy and AUC plots, metrics drop at $\tau=3$ and further at $\tau=4$. This suggests that excessive iterative correction may introduce "over thinking," where the model begins to hallucinate non-existent conflicts or drifts away from the original semantic context, thereby introducing noise rather than clarity. Consequently, we fix $\tau=2$ as the optimal threshold to balance reasoning depth with semantic stability.
\end{itemize}

\indent \textbf{{Summary.}}
Overall, this comprehensive sensitivity analysis validates the robustness of DIVER. While global metrics like Accuracy and AUC exhibit high stability across most hyperparameter variations, the self-correction mechanism demonstrates a distinct pattern where performance peaks at $\tau=2$ before degrading due to over reasoning noise. These results collectively confirm that DIVER achieves a critical balance between reasoning depth and semantic stability, delivering optimal performance when the iterative feedback loop is appropriately regulated.

\section{Appendix C: Computational Cost-Benefit Analysis}
\label{sec:appendix_c}

In this section, we provide a detailed analysis of the computational overhead of DIVER compared to state-of-the-art baselines. As shown in Table \ref{tab:complexity_analysis}, we evaluate efficiency using two key metrics: Average API Calls and Average Token Consumption per sample.

\indent \textbf{Token Efficiency.}
DIVER demonstrates superior efficiency in token usage, achieving the lowest consumption among all methods. Specifically, DIVER reduces the average token usage to 1.5k. Compared to the most efficient baseline (INSIDE with 1.6k tokens), our method reduces token consumption by 0.1k, corresponding to a 6.25\% improvement in token efficiency. This reduction is attributed to our dynamic gating mechanism, which filters out simple samples before they reach the token-heavy forensic reasoning stage.

\indent \textbf{API Call Trade-off.}
Regarding API interactions, DIVER averages 1.3 calls per sample. While this represents a modest increase of 0.3 calls (approximately 30\% overhead) compared to the minimal baseline (INSIDE/GLPN-LLM with 1.0 calls), it remains significantly lower than the exhaustive LIFE method (2.0 calls). We argue that this marginal cost increase is justified by the substantial performance gains: DIVER achieves improvements of 2.20\%--3.50\% in accuracy across datasets. The additional 0.3 calls reflect the selective activation of the visual forensics module only for high-uncertainty samples, ensuring that computational resources are allocated where they contribute most to detection accuracy.

\end{document}